\documentclass[twocolumn]{autart}
\usepackage{graphicx}
\usepackage{caption}
\usepackage{tabularx,booktabs}

\usepackage{multicol}
\usepackage{amsmath,amssymb}
\usepackage{amsfonts}
\usepackage{bbm}
\usepackage{textcomp}
\usepackage{psfrag}
\usepackage{multimedia}
\usepackage{fancybox}

\usepackage[utf8]{inputenc}

\newcommand{\E}[2]{\mathbb{E}_{#1}\left[#2\right]}
\newcommand{\Prob}[2]{\mathbb{P}_{#1}\left[#2\right]}
\newcommand{\tr}{\mathrm{Tr}}
\newtheorem{theorem}{Theorem}
\newtheorem{Proposition}{Proposition}

\newtheorem{Lemma}{Lemma}

\newtheorem{Assumption}{Assumption}

\usepackage[usenames,dvipsnames]{xcolor}
\definecolor{wheat}{rgb}{0.96,0.87,0.70}
\definecolor{mario}{rgb}{0.8,0.8,1}
\definecolor{seb}{rgb}{0.8,1,0.8}

\newcommand {\matr}[2]{\left[\begin{array}{#1}#2\end{array}\right]}
\newcommand {\cmatr}[2]{\left\{\begin{array}{#1}#2\end{array}\right.}
\newcommand{\tightMatr}[2]{\begin{bmatrix}#2\end{bmatrix}}
\newcommand{\norm}[1]{\left \|\begin{matrix}#1\end{matrix}\right \|}
\newcommand{\vect}[1]{{\ensuremath{\boldsymbol{{#1}}}}}

\newcounter{lastnote}

\begin{document} 

\begin{frontmatter}
	
\title{Stability-Constrained Markov Decision Processes Using MPC\thanksref{footnoteinfo}}

\thanks[footnoteinfo]{This paper was not presented at any IFAC 
	meeting. Corresponding author M. Zanon. This paper was partially supported by the Italian Ministry of University and Research under the PRIN'17 project ``Data-driven learning of constrained control systems" , contract no. 2017J89ARP; by ARTES 4.0 Advanced Robotics and enabling digital Technologies \& Systems 4.0, CUP: B81J18000090008; and by the Norwegian Research Council project ``Safe Reinforcement-Learning using MPC" (SARLEM).}

\author[Mario]{Mario Zanon}\ead{mario.zanon@imtlucca.it},
\author[Seb]{S\'ebastien Gros}, 
\author[Michele]{Michele Palladino}, 

\address[Mario]{IMT School for Advanced Studies Lucca, Piazza San Francesco 19, 55100, Lucca, Italy}
\address[Seb]{NTNU, Gløshaugen, Trondheim, Norway}  
\address[Michele]{Gran Sasso Science Institute - GSSI, via Michele Jacobucci 2, 67100, L'Aquila, Italy}



	\begin{abstract}
		In this paper, we consider solving discounted Markov Decision Processes (MDPs) under the constraint that the resulting policy is stabilizing. In practice MDPs are solved based on some form of policy approximation. We will leverage recent results proposing to use Model Predictive Control (MPC) as a structured policy in the context of Reinforcement Learning to make it possible to introduce stability requirements directly inside the MPC-based policy. This will restrict the solution of the MDP to stabilizing policies by construction. The stability theory for MPC is most mature for the undiscounted MPC case. Hence, we will first show in this paper that stable discounted MDPs can be reformulated as undiscounted ones. This observation will entail that the MPC-based policy with stability requirements will produce the optimal policy for the discounted MDP if it is stable, and the best stabilizing policy otherwise.
	\end{abstract}
	
\end{frontmatter}

\section{Introduction}

Markov Decision Processes (MDPs) include a wide class of problems in which a controlled stochastic system needs to minimize a prescribed cost function. A special case is obtained for deterministic systems, in which case the problem is often labelled optimal control. MDPs have been extensively studied~\cite{Bertsekas2007,Bertsekas1996a,Puterman1994,Sutton2018}. Most of the existing literature focuses on studying the theoretical properties of MDPs from an optimization point of view, and on deriving algorithms to solve them, i.e., to compute optimal control policies.

	Solving MDPs exactly is notoriously difficult, and practical approaches often rely on approximate Dynamic Programming or Reinforcement Learning (RL), using some form of function approximation~\cite{Bertsekas1996a,Sutton2018}. The latter approximation approach has recently demonstrated the ability to solve problems that were previously considered intractable, see, e.g.,~\cite{Abbeel2007,Wang2012}. A recently proposed function approximator for RL is Model Predictive Control (MPC)~\cite{Gros2020,Gros2020b,Gros2020a,Zanon2021,Zanon2020,Zanon2019}. 
	One of the benefits of using MPC as a function approximator is the existence of a strong theory proving desirable properties such as safety, stability and some form of explainability~\cite{Grune2011,Rawlings2017}. This fact has motivated recent interest in combining MPC with learning techniques, see, e.g.,~\cite{Aswani2013,Berkenkamp2017,Dean2019,Koller2018,Ostafew2016,Wabersich2019}.

To the best of our knowledge, limited attention has been devoted to the enforcement of stability conditions in MDPs. The study of the stability properties of Markov Chains has been extensively studied in~\cite{Meyn2009}, while the stability properties of undiscounted optimal control for both deterministic systems and stochastic systems with bounded noise have been studied, e.g., in~\cite{Grune2011,Rawlings2017}. The derivation of conditions for the stability of discounted problems is harder and some results have been obtained in, e.g.,~\cite{Gaitsgory2018,Postoyan2014}.

Unfortunately, deriving conditions for stability is in general difficult; moreover, such conditions are often difficult to verify in practice. In this paper we aim at imposing stability conditions by explicitly constraining the candidate policies to be stabilizing. We will do so by leveraging the existing stability theory for undiscounted MPC. As proven in~\cite{Gros2020}, under mild conditions, the MDP optimal policy can be captured via an MPC scheme with an approximate model provided that it is discounted by the same factor as the MDP. 

Because we want to capture the solution of a discounted MDP using an undiscounted MPC, the theoretical gap between the discounted and undiscounted case must be closed first. To that end we will prove that, under a weak stability condition, a given discounted MDP can be formulated as an undiscounted MDP delivering the same optimal policy. Using this result, the MPC-based policy restricted to stability will deliver the optimal policy for the discounted MDP if it is stable, and the optimal policy among the stable policies if it is not.

This paper is structured as follows. In Section~\ref{sec:preliminaries} we will provide the mathematical background and the required definitions. In Section~\ref{sec:Equivalence} we will derive an undiscounted MDP whose optimal policy is also optimal for a given discounted MDP; we will relate the optimality criterion for this undiscounted MDP to commonly used optimality criteria; and we will illustrate the theory in the context of the linear quadratic regulator. We will then formulate the stability constraints in Section~\ref{sec:constraints}, where we will exploit the properties of MPC as a function approximator to solve MDPs. We will provide numerical examples in Section~\ref{sec:simulations} and conclude with Section~\ref{sec:conclusions}.


\section{Preliminaries} 
\label{sec:preliminaries}

We will consider that the system is described by a Markov Decision Process (MDP) having the (possibly) stochastic state transition dynamics 
\begin{equation}
	\Prob{}{\vect{s}_+\,|\, \vect{s},\vect{a}}, \label{eq:TrueDynamics}
\end{equation}
where $\vect{s},\vect{a}$ is the current state-action pair and $\vect{s}_+$ is the subsequent state. We will generally assume that the state-action space is continuous but the theory proposed here is valid in general. Note that notation \eqref{eq:TrueDynamics} is standard in the literature on MDPs, while the control literature typically uses the notation $\vect{s}_+=\vect{f}(\vect{s},\vect{a},\vect{\zeta})$, where $\vect{\zeta}$ is a stochastic variable and $\vect{f}$ a possibly nonlinear function. For discrete state spaces, \eqref{eq:TrueDynamics} is a probability, while for continuous state spaces it is a probability density.

We will label $L(\vect{s},\vect{a})$ the stage cost associated to the MDP, which we will assume can take the form
\begin{equation}
	\label{eq:OriginalStageCost}
	L\left(\vect{s},\vect{a}\right) = l \left(\vect{s},\vect{a}\right) +\mathcal{I}_\infty\left(\vect{h}\left(\vect{s},\vect{a}\right)\right)+ \mathcal{I}_\infty\left(\vect{g}\left(\vect{a}\right)\right),
\end{equation}
where we use the indicator function
\begin{equation}
	\mathcal{I}_\infty(\vect{x}) = \cmatr{lll}{\infty & \ &\text{if }\vect{x}_i >0\text{ for some }i \\ 0& \ &\text{otherwise}}.
\end{equation}
In \eqref{eq:OriginalStageCost}, function $l$ captures the cost given to different state-input pairs, while the constraints
\begin{equation}
	\label{eq:FiniteL}
	\vect{g}(\vect{a}) \leq 0, \qquad \vect{h}(\vect{s},\vect{a})\leq 0,
\end{equation}
capture undesirable state and inputs, and infinite values are given to $L$ when \eqref{eq:FiniteL} is violated. 

\begin{Assumption}
	\label{ass:Basics}
	The cost $l(\vect s,\vect a)$ is finite for all finite states and inputs $\vect s$,$\vect a$. Additionally, for continuous state spaces, \eqref{eq:TrueDynamics} satisfies
	\begin{equation}
	\label{eq:DecayingTails}
	\lim_{\alpha \rightarrow \infty } \Prob{}{\alpha\vect{s}_+\,|\, \vect{s},\vect{a}} = 0,\qquad \forall\, \vect{s}_+, \vect{s},\vect{a}\,\,\text{finite}.
	\end{equation}
\end{Assumption}
The mild assumption~\eqref{eq:DecayingTails} entails that the stage cost $l$ remains bounded over finite horizons with probability 1. This allows us to avoid cumbersome technicalities in the proofs.

In this paper, we aim at solving the stability-constrained discounted MDP
\begin{align}
	\label{eq:stab_constr_disc_mdp}
	\min_{\vect{\pi}\in\Pi_\mathrm{s}} \ \ \E{{\vect{\pi}}}{\sum_{k=0}^{\infty} \gamma^k L(\vect{s}_k,\vect{\pi}(\vect{s}_k)) },
\end{align}
where $\Pi_\mathrm{s}$ denotes the set of all policies $\vect{\pi}$ which stabilize system~\eqref{eq:TrueDynamics}. At this stage, we leave the stability concept in use here intentionally vague, and we will discuss it more in detail in Section~\ref{sec:constraints}.

Since in this paper we will compare discounted MDPs with undiscounted MDPs, we summarize next the corresponding optimality concepts. All definitions we will provide are given without stability constraints, but we remark that the same definitions hold also if the additional constraint $\vect{\pi}\in\Pi_\mathrm{s}$ is introduced. 
For more details on MDPs and optimality notions we refer to, e.g.,~\cite{Puterman1994} and references therein.

\subsection{Optimality of Discounted MDPs}

With the introduction of a discount factor $0<\gamma \leq 1$, given \eqref{eq:TrueDynamics} and \eqref{eq:OriginalStageCost}, the optimal discounted policy $\vect{\pi}_\star^\gamma $ is the policy minimizing the expected total discounted cost
\begin{equation}
	\label{eq:discuonted_optimal_policy}
	\vect{\pi}_\star^\gamma = \arg \min_{\vect{\pi}} \ \ \E{{\vect{\pi}}}{\sum_{k=0}^{\infty} \gamma^k L(\vect{s}_k,\vect{\pi}(\vect{s}_k)) }.
\end{equation}
where the expected value $ \E{{\vect{\pi}}}{\cdot}$ is taken over the state transition dynamics \eqref{eq:TrueDynamics} in closed loop with policy $\vect{\pi}$.  
For any policy we define the associated action-value function $Q_{\vect{\pi}}^\gamma $ and value function $V_{\vect{\pi}}^\gamma $ as
\begin{subequations}
	\label{eq:disc_value_functions}
	\begin{align}
		V_{\vect{\pi}}^\gamma(\vect{s}) &:=\E{{\vect{\pi}}}{\left.\sum_{k=0}^{\infty} \gamma^k L(\vect{s}_k,\vect{\pi}(\vect{s}_k)) \, \right|\, \vect{s}_0 =\vect{s} }, \\
		Q_{\vect{\pi}}^\gamma(\vect{s},\vect{a}) &:= L(\vect{s},\vect{a}) + \gamma \E{}{V_{\vect{\pi}}^\gamma(\vect{s}_+) \, |\, \vect{s}, \vect{a} }.
	\end{align}
\end{subequations}
The optimal action-value function $Q_\star^\gamma =Q_{\vect{\pi}_\star^\gamma}^\gamma$ and value function $V_\star^\gamma =V_{\vect{\pi}_\star^\gamma}^\gamma$ associated with the discounted MDP 
are defined by the Bellman equations~\cite{Bertsekas2005}:
\begin{subequations}
\label{eq:Bellman}
\begin{align}
Q_\star^\gamma \left(\vect{s},\vect{a}\right) &= L\left(\vect{s},\vect{a}\right) + \gamma \mathbb{E}{}\left[V_\star^\gamma (\vect{s}_+)\,|\, \vect{s},\vect{a}\right], \label{eq:Bellman1}\\
V_\star^\gamma \left(\vect{s}\right) &= Q_\star^\gamma \left(\vect{s},\vect{\pi}_\star^\gamma \left(\vect{s}\right)\right) = \min_{\vect{a}}\, Q_\star^\gamma \left(\vect{s},\vect{a}\right). \label{eq:Bellman2} 
\end{align}
\end{subequations}
Throughout the paper we will assume that the MDP underlying the system, the associated stage cost $L$ and the discount factor $\gamma$ yields a well-posed problem, i.e., the value functions defined by \eqref{eq:Bellman} are well-posed, and finite over some non-empty sets. This well-posedness is formulated in the following assumption.
\begin{Assumption}
	\label{ass:Vstability}
	There exists a nonempty set $\mathcal{S}$ such that for all $\vect{s}\in\mathcal{S}$ it holds that
	\begin{equation}
		\label{eq:ass_stability}
		|V_\star^\gamma (\vect{ s})|<\infty.
	\end{equation}
\end{Assumption}

We state next an immediate consequence of this assumption, which will be useful afterwards.
\begin{Lemma} \label{Lem:FiniteV}
	Suppose that Assumptions~\ref{ass:Basics}-\ref{ass:Vstability} hold. Then,
	\begin{equation}
	\label{eq:V_stability}
	-\infty <  \E{\vect{\pi}_\star}{V_\star^\gamma (\vect{ s}_k)\,|\, \vect s_0 = \vect s}  <  \infty, \ \ \forall \ k = 0,\ldots,N,
	\end{equation}
	holds for all $ \vect s \in \mathcal S$, and $N$ finite.
\end{Lemma}
\begin{pf}
	We observe that
	\begin{align}
	\label{eq:ConnectFiniteV}
	&\E{\vect{\pi}_\star}{V_\star^\gamma (\vect{ s}_k)\,|\, \vect s_0} = \\ 
	&\qquad\gamma^{-k}V_\star^\gamma (\vect{ s}_0)-\gamma^{-k}\,\E{\vect{\pi}_\star}{\left.\sum_{i=0}^{k-1}\gamma^iL\left(\vect s_i,\vect\pi_\star\left(\vect s_i\right)\right)\,\right|\, \vect s_0}.  \nonumber
	\end{align}
	Since $ \gamma^{-k}V_\star^\gamma (\vect{ s}_0)$ is bounded for any $k< \infty$ and $\vect s_0 \in \mathcal S$, due to Assumption~\ref{ass:Basics} and Equation~\eqref{eq:DecayingTails}, the second term of the right-hand side of \eqref{eq:ConnectFiniteV} must be lower bounded for all $k$ finite. Additionally, in order for $V_\star^\gamma (\vect{ s}_0)$ to be finite, it must also be upper bounded. Consequently, $\E{\vect{\pi}_\star}{V_\star^\gamma (\vect{ s}_k)\,|\, \vect s_0} $ must be bounded.
	$\hfill\qed$
\end{pf}

\subsection{Optimality of Undiscounted MDPs}

For undiscounted MDPs, a discount factor $\gamma = 1$ is selected, such that the cost in~\eqref{eq:discuonted_optimal_policy} can be unbounded. Optimal policies are then typically defined according to a hierarchy of criteria. The first criterion is based on the expected average total cost
\begin{equation}
	\label{eq:gain_optimal:policy}
	\vect{\bar \pi}_\star = \arg \min_{\vect{\pi}} \  \lim_{N\to\infty} \frac{1}{N} \E{{\vect{\pi}}}{\sum_{k=0}^{N-1} L(\vect{s}_k,\vect{\pi}(\vect{s}_k)) },
\end{equation}
with \emph{gain}, or average cost
\begin{equation}
	\label{eq:gain_optimal:averagecost}
	\bar L_\infty\left(\vect s_0\right) = \lim_{N\to\infty} \frac{1}{N}\E{\vect{\bar \pi}_\star\hspace{-3pt}}{\left.\sum_{k=0}^{N-1} L(\vect{s}_k,\vect{\bar \pi}_\star(\vect{s}_k)) \,\right| \vect s_0}.
\end{equation}
Any policy satisfying~\eqref{eq:gain_optimal:policy} is called \emph{gain optimal}. The average cost $\bar L_\infty$ is often assumed to be independent of the initial state $\vect s_0$ and we will also make this assumption for the sake of simplicity. For more information on the general case, we refer the interested reader to~\cite{Puterman1994}. We ought to stress here that gain optimal policies are in general not unique.

The gain optimality criterion only ensures that the optimal steady-state distribution of the closed-loop Markov chain is asymptotically reached, but it disregards transients. The concept of \emph{bias optimality} has been introduced to account for the optimality of transients.
A \emph{bias optimal} policy $\vect{\tilde \pi}_\star$ minimizes the total undiscounted cost
\begin{equation}
	\label{eq:bias_optimal_definition}
	\vect{\tilde \pi}_\star  = \arg \min_{\vect{\pi}} \  \E{{\vect{\pi}}}{\sum_{k=0}^{\infty} \left (L(\vect{s}_k,\vect{\pi}(\vect{s}_k)) -  \bar L_\infty\right )}.
\end{equation}

Optimality notions which are more stringent than bias optimality are beyond the scope of this paper. We simply recall that bias optimal policies are gain optimal and that the most stringent optimality criterion for undiscounted MDPs is \emph{Blackwell optimality}~\cite{Blackwell1962,Puterman1994}.

\section{Equivalent Undiscounted MDP Formulation}
\label{sec:Equivalence}

In this section, we discuss the equivalence between discounted MDPs and suitably formulated undiscounted MDPs. To that end, let us define the modified stage cost
\begin{equation}
	\label{eq:Ltilde:def}
	\tilde L^\gamma(\vect{s},\vect{a}) := L(\vect{s},\vect{a}) + (\gamma -1)\E{}{V_\star^\gamma (\vect{ s}_+) | \vect{s}, \vect{a}},
\end{equation}
where we explicitly state the dependence on the discount factor $\gamma$ to stress the fact that its definition is based on a discounted MDP formulation.

In the following, we construct the theory allowing one to support the discounted MDP solution by using undiscounted, finite-horizon stochastic MPC schemes. The latter are themselves undiscounted finite-horizon MDPs. Hence, it will be useful to first  connect infinite-horizon discounted MDPs to finite-horizon undiscounted ones. We establish the equivalence between infinite-horizon discounted and infinite-horizon undiscounted MDPs later in the text.

In order to build these equivalences, let us define the $N$-steps undiscounted value and action-value functions as
\begin{subequations}
\label{eq:undisc_value_functions_Nsteps}
	\begin{align}
		&\tilde V_{\vect{\pi}}^{\gamma,N}(\vect{s}) \hspace{-1pt} := \hspace{-1pt} \E{\vect{\pi}\hspace{-3pt}}{\left.  V_\star^\gamma (\vect{ s}_N) +\hspace{-1pt} \sum_{k=0}^{N-1} \tilde L^\gamma(\vect{s}_k,\vect{\pi}(\vect{s}_k)) \, \right|  \vect{s}_0 = \vect{s}  }\hspace{-1pt}, \label{eq:Vtilde}\\
		&\tilde Q_{\vect{\pi}}^{\gamma,N}(\vect{s},\vect{a}) 
		\hspace{-1pt}:=\hspace{-1pt}\tilde L^\gamma(\vect{s},\vect{a}) + \E{\hspace{-1pt}}{\left.  \tilde V_{\vect{\pi}}^{\gamma,N-1}(\vect{s}_+) \, \right|  \vect{s}, \vect{a}  }\hspace{-1pt} . \label{eq:Qtilde}
	\end{align}
\end{subequations}
We define the corresponding optimal policy as 
\begin{equation}
\label{eq:optimal_undisc_value_functions_Nsteps}
\vect{\tilde \pi}_\star^{\gamma,N} := \mathrm{arg}\min_{\vect\pi}\, \tilde V_{\vect{\pi}}^{\gamma,N}(\vect{s}), 
\end{equation}
with associated optimal value and action-value function $\tilde V_{\star}^{\gamma,N}:=\tilde V_{\vect{\tilde \pi}_\star^{\gamma,N}}^{\gamma,N}$, and $\tilde Q_{\star}^{\gamma,N}:=\tilde Q_{\vect{\tilde \pi}_\star^{\gamma,N}}^{\gamma,N}$.
We will justify in Section~\ref{sec:optimality} the use of the $\tilde \cdot$ notation, which we also used to define bias-optimal quantities. 
We deliver next our first result, which proves the equivalence between the optimal undiscounted $N$-steps value functions and the optimal discounted value functions.

%

\begin{theorem}
	\label{thm:undiscounted}
	Suppose that Assumption~\ref{ass:Vstability} holds for all $\vect{s}\in\mathcal{S}$. 
	Then $\forall\,\vect s\in\mathcal{S}$, $\forall\,\vect a,\,N <\infty$ it holds that 
	\begin{equation*}
	\tilde V_\star^{\gamma,N}(\vect{s}) = V_\star^\gamma (\vect{s}), \qquad
	\tilde Q_\star^{\gamma,N}(\vect{s},\vect{a}) = Q_\star^\gamma (\vect{s},\vect{a}).
	\end{equation*}
\end{theorem}
\begin{pf}
	We use the Bellman Equation~\eqref{eq:Bellman} to obtain
	\begin{align}
		\label{eq:L_V}
		L(\vect{s},\vect{\pi}_\star^\gamma(\vect{s})) = V_\star^\gamma (\vect{s}) - \gamma \E{}{V_\star^\gamma (\vect{ s}_+) | \vect{s}, \vect{\pi}_\star^\gamma(\vect{s})},
	\end{align}
	which we use together with~\eqref{eq:Ltilde:def} to obtain
	\begin{equation}
		\label{eq:Ltilde_Vtilde}
		\tilde L^\gamma(\vect{s},\vect{\pi}_\star^\gamma(\vect{s})) = V_\star^\gamma (\vect{ s}) - \E{}{V_\star^\gamma (\vect{ s}_+) | \vect{s}, \vect{a}}.
	\end{equation}
	Equation~\eqref{eq:Vtilde} then becomes the telescopic sum:
	\begin{equation*}
		\tilde V_{\vect{\pi}_\star^\gamma}^{\gamma,N}(\vect{s}_0) = \E{\vect{\pi}_\star^\gamma}{ V_\star^\gamma (\vect{ s}_N) + \sum_{k=0}^{N-1} V_\star^\gamma (\vect{ s}_k) - V_\star^\gamma (\vect{ s}_{k+1})   }.
	\end{equation*}
	Using Assumption~\ref{ass:Vstability}, we simplify the terms in the telescopic sum to obtain
	\begin{equation}
		\label{eq:VtildeVstar}
		\tilde V_{\vect{\pi}_\star^\gamma}^{\gamma,\bar N}(\vect{s}) = V_\star^\gamma (\vect{s}), \qquad \forall \ \bar N \leq N.
	\end{equation}
	Consequently, by~\eqref{eq:Qtilde} we have
	\begin{align*}
	\tilde Q^{\gamma,N}_{\vect{\pi}_\star^\gamma}(\vect{s},\vect{a}) &= \tilde L^\gamma(\vect{s},\vect{a}) + \E{}{\tilde V_{\vect{\pi}_\star^\gamma }^{\gamma,N-1}(\vect{ s}_+) | \vect{s}, \vect{a}} \\
	&= L(\vect{s},\vect{a}) + \gamma \E{}{ V_{\star}^\gamma (\vect{ s}_+) | \vect{s}, \vect{a}} = Q_\star^\gamma (\vect{s},\vect{a}),
	\end{align*}
	where we exploited~\eqref{eq:Ltilde:def} and~\eqref{eq:VtildeVstar}.
	Since
	\begin{equation*}
	\vect{\pi}_\star^\gamma  = \arg\min_{\vect{a}} \, Q_\star^\gamma \left(\vect{s},\vect{a}\right) = \arg\min_{\vect{a}} \, \tilde Q_{\vect{\pi}_\star^\gamma}^{\gamma,N}\left(\vect{s},\vect{a}\right)=\vect{\tilde \pi}_\star^{\gamma,N},
	\end{equation*}
	we immediately obtain
	\begin{align*}
	\tilde V_\star^{\gamma,N}(\vect{s}) &= \tilde V_{\vect{\tilde \pi}^{\gamma,N}_\star }^{\gamma,N}(\vect{s}) = \tilde V_{\vect{\pi}_\star^\gamma  }^{\gamma,N}(\vect{s})=  V_\star^\gamma (\vect{s}), \\ 
	\tilde Q_\star^{\gamma,N}(\vect{s},\vect{a}) &= \tilde Q_{\vect{\tilde \pi}^{\gamma,N}_\star}^{\gamma,N}(\vect{s},\vect{a}) = \tilde Q_{\vect{\pi}_\star^\gamma }^{\gamma,N}(\vect{s},\vect{a})= Q_\star^\gamma (\vect{s},\vect{a}).
	\end{align*}
	$\hfill\qed$
\end{pf}

If the value function $\tilde V_\star^{\gamma,N}$ remains bounded for  all $\vect s\in\mathcal{S}$ as $N\to\infty$, then the result above holds also as $N\to\infty$, but the terminal cost $V_\star^\gamma$ is  still required in forming $\tilde V_{\vect{\pi}}^{\gamma,N}$, $\tilde Q_{\vect{\pi}}^{\gamma,N}$. In order to dismiss that terminal cost for $N\rightarrow\infty$ we need an additional stronger assumption. 
\begin{Assumption}
	\label{ass:Vstability_inf}
	Assumption~\ref{ass:Vstability} holds. Moreover,
	\begin{equation}
	\label{eq:Stab:Condition}
		-\infty < \lim_{k\to\infty} \ \E{\vect{\pi}_\star^\gamma}{V_\star^\gamma (\vect{ s}_k)\,|\, \vect s_0 } := v_\infty^\gamma < \infty,
	\end{equation}
	holds $\forall\,\vect s_0\in\mathcal S$ for some constant $v_\infty^\gamma$. 
	\end{Assumption}

Assumption~\ref{ass:Vstability_inf} entails a weak form of stability for the discounted MDP, as we discuss next. It is stronger than requiring the existence of a bounded value function, i.e.,
\begin{equation}
\label{eq:V:existence}
\left |\lim_{N\to\infty} \ \E{\vect{\pi}_\star^\gamma }{ \left.\sum_{k=0}^{N-1} \gamma^k L (\vect{ s}_k,\vect \pi^\gamma_\star(\vect s_k))\,\right |\, \vect s_0}\right | \leq \infty,
\end{equation}
holds on a non-empty set of initial conditions $\vect s_0$. Indeed, \eqref{eq:V:existence} is finite provided that the stage cost $L (\vect{ s}_k,\vect \pi^\gamma_\star(\vect s_k))$ diverges in expectation  at a rate no larger than $\gamma_\mathrm D^{-k}$ for some $\gamma_\mathrm D > \gamma$. It follows that \eqref{eq:V:existence} allows the cost to grow unbounded over time. Hence, the existence of a bounded value function (Assumption~\ref{ass:Vstability}) does not entail that the value function remains bounded over time. 
Moreover, assuming~\eqref{eq:Stab:Condition}, i.e., that the limit exists and converges to the constant $v_\infty^\gamma$, introduces further restrictions, which rule out, e.g., periodic oscillations of the cost.
We cast Assumption~\ref{ass:Vstability_inf} as a weak form of stability: if the MDP converges to a unique steady-state distribution, then Assumption~\ref{ass:Vstability_inf} automatically holds. However, converging to a steady-state distribution is a stronger requirement, since Assumption~\ref{ass:Vstability_inf} might hold also for non-steady-state, and even diverging distributions, see, e.g., Section~\ref{sec:lqr}.


%

In order to formulate the next theorem, let us use \eqref{eq:Ltilde:def} to first define the undiscounted value and action-value functions without terminal cost as
\begin{subequations}
	\label{eq:undisc_value_functions}
	\begin{align}
		&\tilde V^\gamma_{\vect{\pi}}(\vect{s}) := \lim_{N\to\infty} \hspace{-1pt}\E{\vect{\pi}\hspace{-3pt}}{  \left. \sum_{k=0}^{N-1} \hspace{-2pt}\tilde L^\gamma(\vect{s}_k,\vect{\pi}(\vect{s}_k)) \, \right|  \vect{s}_0 = \vect{s}  }, \label{eq:Vtilde_inf}\\
		&\tilde Q^\gamma_{\vect{\pi}}(\vect{s},\vect{a}) :=\tilde L^\gamma(\vect{s},\vect{a}) + \E{}{ \left. \tilde V^\gamma_{\vect{\pi}}(\vect{s}_+) \, \right|  \vect{s}, \vect{a}  }, \label{eq:Qtilde_inf}
	\end{align}
\end{subequations}
which do not necessarily match the limit for $N\to\infty$ of the value functions defined in~\eqref{eq:undisc_value_functions_Nsteps}. Furthermore, we define the optimal undiscounted policy and the associated value functions as:
\begin{subequations}
\label{eq:optimal_undiscounted_policy:Gen}
	\begin{align}
		\label{eq:optimal_undiscounted_policy}
		&\vect{\tilde \pi}_\star^\gamma := \mathrm{arg}\min_{\vect\pi}\,  \lim_{N\to\infty} \E{\vect{\pi}}{ \ \sum_{k=0}^{N-1} \tilde L^\gamma(\vect{s}_k,\vect{\pi}(\vect{s}_k)) \   }, \\
		&\tilde V^\gamma_{\star}(\vect{s}) := \tilde V^\gamma_{\vect{\tilde \pi}_\star^\gamma}(\vect{s}), \qquad \tilde Q^\gamma_{\star}(\vect{s},\vect a) := \tilde Q^\gamma_{\vect{\tilde \pi}_\star^\gamma}(\vect{s},\vect a).
	\end{align}
\end{subequations}

\begin{theorem}
	\label{thm:undiscounted_inf_hor}
	Suppose that Assumption~\ref{ass:Vstability_inf} holds. Then $\forall\,\vect s\in\mathcal{S}$, $\forall\,\vect a$ it holds that 
	\begin{subequations}
	\begin{align}
	&\hspace{-5pt}\tilde V^\gamma_\star(\vect{s}) \hspace{-1pt}=\hspace{-1pt} V_\star^\gamma (\vect{s}) \hspace{-1pt}-\hspace{-1pt} v_\infty^\gamma, \label{eq:valueFunctionEquality} \ \
	\tilde Q^\gamma_\star(\vect{s},\vect{a}) \hspace{-1pt}=\hspace{-1pt} Q_\star^\gamma (\vect{s},\vect{a})\hspace{-1pt} -\hspace{-1pt} v_\infty^\gamma, \\
	&\hspace{-5pt}\vect{\tilde \pi}_\star^\gamma\left(\vect s\right) \hspace{-1pt}= \vect\pi_\star^\gamma \left(\vect s\right). \label{eq:policyEquality}
	\end{align}
	\end{subequations}
\end{theorem}
\begin{pf}
		Similar to the proof of Theorem~\ref{thm:undiscounted}, we use the Bellman Equation~\eqref{eq:Bellman} together with~\eqref{eq:L_V}-\eqref{eq:Ltilde_Vtilde} to write~\eqref{eq:Vtilde_inf} as a telescopic sum in which all terms are bounded due to Lemma \ref{Lem:FiniteV}. By simplifying the terms in the sum we obtain
		\begin{equation}
			\label{eq:Vtilde_infVstar}
			\tilde V^\gamma_{\vect{\pi}_\star^\gamma}(\vect{s}) \hspace{-1pt}=\hspace{-1pt} V_\star^\gamma (\vect{s}) -\hspace{-1pt} \lim_{k\to \infty} \E{\vect{\pi}_\star^\gamma\hspace{-2pt} }{V_\star^\gamma (\vect{ s}_k)} \hspace{-1pt}=\hspace{-1pt} V_\star^\gamma (\vect{s}) - v_\infty^\gamma.
		\end{equation}
		Consequently, 
		\begin{align*}
			\tilde Q^\gamma_{\vect{\pi}_\star^\gamma}(\vect{s},\vect{a}) &= \tilde L^\gamma(\vect{s},\vect{a}) + \E{}{\tilde V^\gamma_{\vect{\pi}_\star^\gamma}(\vect{ s}_+) | \vect{s}, \vect{a}} \\
			&= L(\vect{s},\vect{a}) + \gamma \E{}{ V_{\star}^\gamma (\vect{ s}_+) | \vect{s}, \vect{a}} - v_\infty^\gamma \\
			&= Q_\star^\gamma (\vect{s},\vect{a}) - v_\infty^\gamma,
		\end{align*}
		where we used~\eqref{eq:Ltilde:def} and~\eqref{eq:Vtilde_infVstar}.
		Then,
		\begin{equation}
			\vect{\tilde \pi}_\star^\gamma \hspace{-1pt}=\hspace{-1pt} \arg\min_{\vect{a}} \, \tilde Q_{\vect{\pi}_\star^\gamma}^\gamma\hspace{-1pt} \left(\vect{s},\vect{a}\right) \hspace{-1pt}=\hspace{-1pt} \arg\min_{\vect{a}} \, Q^\gamma_{\star}\hspace{-3pt}\left(\vect{s},\vect{a}\right) \hspace{-1pt}=\hspace{-1pt} \vect{ \pi}^\gamma_\star,
		\end{equation}
		which immediately entails~\eqref{eq:policyEquality} and, in turn,
		\begin{align*}
		\tilde V^\gamma_\star(\vect{s}) &= \tilde V^\gamma_{\vect{\tilde \pi}^\gamma_\star}(\vect{s}) = \tilde V^\gamma_{\vect{\pi}_\star^\gamma }(\vect{s}) = V_\star^\gamma (\vect{s})- v_\infty^\gamma, \\ 
		\tilde Q^\gamma_\star(\vect{s},\vect{a}) &= \tilde Q^\gamma_{\vect{\tilde \pi}^\gamma_\star}(\vect{s},\vect{a}) = \tilde Q^\gamma_{\vect{\pi}_\star^\gamma }(\vect{s},\vect{a}) = Q_\star^\gamma (\vect{s},\vect{a})- v_\infty^\gamma.
		\end{align*}
	$\hfill\qed$
\end{pf}

This theorem establishes that any discounted MDP satisfying Assumption~\ref{ass:Vstability_inf} can be reformulated as an undiscounted MDP which delivers the same policy and the same value functions up to a constant term. 

We ought to stress here that in the following we are interested in forming policies which are stabilizing by construction, i.e., we aim at solving the discounted MDP under the constraint of preserving stability. The equivalence proposed in Theorem~\ref{thm:undiscounted_inf_hor} will be instrumental in allowing us to formulate such constraints in the undiscounted setting, while optimizing the cost in a discounted sense. 
This is of particular interest, since the stability analysis is much simpler and more developed for the undiscounted setting. We will discuss the introduction of stability constraints in Section~\ref{sec:constraints}. 
Before detailing how one can introduce stability constraints, we first prove that the obtained undiscounted MDP \eqref{eq:undisc_value_functions}-\eqref{eq:optimal_undiscounted_policy:Gen} yields bias optimal policies; and we then illustrate the theoretical developments in the simple case of a Linear Quadratic Regulator (LQR).

\subsection{Equivalence of Optimality Notions}
\label{sec:optimality}

The undiscounted MDP used in Theorem~\ref{thm:undiscounted_inf_hor} minimizes the cost in~\eqref{eq:optimal_undiscounted_policy}, which is not directly related to standard optimality concepts such as gain or bias optimality. We therefore prove next that the policy $\vect{\tilde \pi}_\star^\gamma=\vect{\pi}_\star^\gamma$ obtained from~\eqref{eq:optimal_undiscounted_policy} is in fact bias optimal. To that end, we will first prove gain optimality. We will then prove that the optimal gain is $0$, which we will relate to bias optimality.

\begin{theorem}
	\label{thm:gain_optimal}
	Suppose that Assumption~\ref{ass:Vstability_inf} holds. Then, policy $\vect{\pi}_\star^\gamma$ is gain optimal for stage cost $\tilde L^\gamma$. 
\end{theorem}
\begin{pf}
	By Theorem~\ref{thm:undiscounted_inf_hor} policy $\vect{\pi}_\star^\gamma$ 
	solves~\eqref{eq:optimal_undiscounted_policy} with a finite optimal cost, such that $\forall \, \vect{\pi}$ we have
	\begin{equation}
		\E{\vect{\pi}_\star^\gamma\hspace{-3pt}}{  \sum_{k=0}^{\infty} \hspace{-1pt}\tilde L^\gamma(\vect{s}_k,\vect{\pi}(\vect{s}_k))   } \hspace{-3pt} \leq \hspace{-1pt}\E{\vect{\pi}\hspace{-3pt}}{  \sum_{k=0}^{\infty} \hspace{-1pt}\tilde L^\gamma(\vect{s}_k,\vect{\pi}(\vect{s}_k))  }\hspace{-3pt}.
	\end{equation}
	Therefore, the policy is gain optimal, i.e., $\forall \, \vect{\pi}$ we have
	\begin{align*}
	&\lim_{N\to\infty} \frac{1}{N} \E{\vect{\pi}_\star^\gamma}{ \ \sum_{k=0}^{N-1} \tilde L^\gamma(\vect{s}_k,\vect{\pi}(\vect{s}_k)) \   } \\
	&\hspace{7em}\leq \lim_{N\to\infty} \frac{1}{N} \E{\vect{\pi}}{ \ \sum_{k=0}^{N-1} \tilde L^\gamma(\vect{s}_k,\vect{\pi}(\vect{s}_k)) \   }.
	\end{align*}
	$\hfill\qed$
\end{pf}

\begin{theorem}
	\label{thm:bias_optimal}
	Suppose that Assumption~\ref{ass:Vstability_inf} holds. Then, policy $\vect{\pi}_\star^\gamma$ is bias optimal for stage cost $\tilde L^\gamma$. 
\end{theorem}
\begin{pf}
	Because $\tilde V_\star^\gamma(\vect{s}_0)$ is finite for all $\vect{s}_0\in\mathcal{S}$, we have that the average cost is
	\begin{align}
	\label{eq:undiscounted_gain_optimal}
	\bar{\tilde L}^\gamma_\infty &= \lim_{N\to\infty} \frac{1}{N}\E{}{\sum_{k=0}^{N-1} \tilde L^\gamma (\vect{s}_k,\vect{\tilde \pi}_\star^\gamma(\vect{s}_k)) } \nonumber\\
	&= \lim_{N\to\infty} \frac{1}{N} \tilde V_\star^\gamma(\vect{s}_0) =0.
	\end{align}
	By Theorem~\ref{thm:gain_optimal} policy $\vect{\tilde \pi}_\star^\gamma=\vect{\pi}_\star^\gamma$ is gain optimal, such that by~\eqref{eq:undiscounted_gain_optimal} the gain-optimal average cost is $\bar{\tilde L}^\gamma_\infty=0$. Therefore, bias optimality for $\tilde L^\gamma$ is obtained by minimizing
	\begin{equation}
	\label{eq:bias_optimality_cost}
	\E{{\vect{\pi}}\hspace{-3pt}}{\sum_{k=0}^{\infty} \tilde L^\gamma(\vect{s}_k,\vect{\pi}(\vect{s}_k)) -  \bar{\tilde L}^\gamma_\infty}\hspace{-2pt}=\E{{\vect{\pi}}\hspace{-3pt}}{\sum_{k=0}^{\infty} \tilde L^\gamma(\vect{s}_k,\vect{\pi}(\vect{s}_k)) },
	\end{equation}
	which is the total undiscounted reward. Since by Theorem~\ref{thm:undiscounted_inf_hor} the policy $\vect{\tilde \pi}_\star^\gamma=\vect{\pi}_\star^\gamma$ solves~\eqref{eq:optimal_undiscounted_policy}, i.e., minimizes~\eqref{eq:bias_optimality_cost}, it is bias optimal by construction.
	$\hfill\qed$
\end{pf}



\subsection{The LQR Case}
\label{sec:lqr}

In order to clarify the previous developments, let us consider the case of a stochastic linear system
\begin{equation}
	\label{eq:linear_mdp}
	\vect{s}_+ = A\vect{s} + B\vect{a} + \vect{w},
\end{equation}
with $\vect{w}\sim \mathcal{N}(\vect{0},W)$  i.i.d., $\E{}{\vect{w}\vect{s}^\top}=0$, $\E{}{\vect{w}\vect{a}^\top}=0$, 
\begin{equation*}
	L(\vect{s},\vect{a}) = \tightMatr{c}{\vect{s} \\ \vect{a}}^\top H \tightMatr{c}{\vect{s} \\ \vect{a}}, \qquad H=\tightMatr{ll}{T & U^\top \\ U & R} \succ0.
\end{equation*}
The value and action-value function are given by
\begin{align*}
	V_\star^\gamma (\vect{s}) &= \vect{s}^\top P \vect{s} + V_0, \qquad\qquad V_0=\frac{\gamma}{1-\gamma}\tr \left ( PW \right ), \\
	Q_\star^\gamma (\vect{s},\vect{a}) \hspace{-1pt}&=\hspace{-2pt} \tightMatr{c}{\hspace{-1pt}\vect{s} \\ \hspace{-1pt}\vect{a}}^{\hspace{-2pt}\top} \hspace{-3pt} \tightMatr{ll}{T \hspace{-1pt}+\hspace{-1pt} \gamma A^\top PA & \hspace{2pt}U^\top  \hspace{-3pt}+\hspace{-1pt} \gamma A^\top PB \\ U \hspace{-1pt}+\hspace{-1pt} \gamma B^\top PA & R \hspace{-1pt}+\hspace{-1pt} \gamma B^\top PB} \hspace{-3pt}\tightMatr{c}{\hspace{-1pt}\vect{s} \\ \hspace{-1pt}\vect{a}} \hspace{-2pt}+\hspace{-1pt} V_0,
\end{align*}
with
\begin{subequations}
	\begin{align}
		\label{eq:satb_dare}
		P &= T + \gamma A^\top P A - (U^\top+\gamma A^\top PB)K,\\
		K &= (R+\gamma B^\top PB)^{-1}(U+\gamma B^\top PA), \\
		0 &\prec R+\gamma B^\top PB. \label{eq:existence_condition}
	\end{align}
\end{subequations}
Note that 
$P$ can be finite even in case $\rho( A-BK ) \geq 1$, where $\rho(\cdot)$ denotes the spectral radius, in which case the value function is defined and bounded for bounded states, but Assumption~\ref{ass:Vstability_inf} does not hold.

\paragraph*{Checking Assumption~\ref{ass:Vstability_inf}}
\label{sec:stoch_lqr_ass2}
We observe that, under feedback $\vect{a}=-K\vect{s}$ we have
\begin{equation*}
	S_+ = A_K S A_K^\top + W, 
\end{equation*}
where we used $A_K:=A-BK$, $S:=\E{}{\vect{s}\vect{s}^\top}$.
If $\rho( A_K ) < 1$, then there exists a unique matrix $S_\infty$ solving the Lyapunov equation $A_K S_\infty A_K^\top - S_\infty + W=0$. 
Assume that $P$ is full rank (which is typically the case), then we have 
\begin{align*}
	\lim_{k\to \infty} \hspace{-1pt}\E{\vect{\pi}_\star\hspace{-2pt}}{V_\star^\gamma (\vect{s}_k)} \hspace{-2pt}=\hspace{-1pt} \left \{ 
	\begin{array}{ll}
	\hspace{-2pt}\infty & \text{if } \rho( A_K ) \hspace{-2pt}\geq\hspace{-1pt} 1 \\
	\hspace{-2pt}\tr \left (PS_\infty\right ) \hspace{-1pt}+\hspace{-1pt} V_0 \hspace{-1pt}< \hspace{-1pt}\infty & \text{otherwise}
	\end{array}
	\right. .
\end{align*}
The condition $\rho(A_K)<1$ distinguishes the case $\lim_{k\to\infty}\E{}{\vect{s}_k}=0$, $\lim_{k\to\infty}\E{}{\vect{s}_k\vect{s}_k^\top}=S_\infty$ from the case $\lim_{k\to\infty}\E{}{\vect{s}_k}=\pm\infty$: in the former, convergence in expectation of the value function is guaranteed; in the latter we immediately have $\lim_{k\to \infty} \E{\vect{\pi}_\star}{V_\star^\gamma (\vect{s}_k)} =\infty$. 

\paragraph*{Undiscounted Equivalent MDP}
We observe that 
\begin{align*}
	\hspace{2em}&\hspace{-2em}\E{}{V_\star^\gamma (\vect{s}_+)|\vect{s},\vect{a}} \\
	&=\tightMatr{c}{\vect{s} \\ \vect{a}}^\top \hspace{-1pt} \tightMatr{ll}{A^\top PA & A^\top PB \\  B^\top PA & B^\top PB} \tightMatr{c}{\vect{s} \\ \vect{a}} + \frac{1}{1-\gamma} \tr\left ( PW \right ).
\end{align*}
In case Assumption~\ref{ass:Vstability_inf} holds, the stage cost for the corresponding undiscounted MDP is then given by
\begin{align}
	\label{eq:tildeL:lqr}
	\tilde L^\gamma(\vect{s},\vect{a}) &= \tightMatr{c}{\vect{s} \\ \vect{a}}^\top \hspace{-1pt} \tilde H \tightMatr{c}{\vect{s} \\ \vect{a}} - \tr \left (PW\right ), \\
	\tilde H&=\tightMatr{ll}{T + (\gamma-1) A^\top PA & U^\top + (\gamma-1) A^\top PB \\ U + (\gamma-1) B^\top PA & R + (\gamma-1) B^\top PB}. \nonumber
\end{align}
Consequently, we have
\begin{align*}
	\tilde V_\star^\gamma(\vect{s}) &\hspace{-1pt}=\hspace{-1pt} V_\star^\gamma (\vect{s}) \hspace{-1pt}-\hspace{-1pt} \tr \left (PS_\infty\right ) \hspace{-1pt}-\hspace{-1pt} V_0\hspace{-1pt}=\hspace{-1pt}\vect{s}^\top \hspace{-1pt} P \vect{s}\hspace{-1pt} -\hspace{-1pt} \tr \left (PS_\infty\right )\hspace{-1pt}, \\
	\tilde Q_\star^\gamma(\vect{s},\vect{a}) &\hspace{-1pt}=\hspace{-1pt} Q_\star^\gamma (\vect{s},\vect{a}) - \tr \left (PS_\infty\right ) - V_0.
\end{align*}
Then, if $P$ is full rank, the following holds:
\begin{align*}
\lim_{k\to \infty} \E{\vect{\pi}_\star}{\tilde V_\star^\gamma(\vect{s}_k)} = \left \{ 
\begin{array}{ll}
\infty & \text{if } \rho( A-BK ) \geq 1 \\
0 & \text{otherwise}
\end{array}
\right. .
\end{align*}
In case $P=0$, Assumption~\ref{ass:Vstability_inf} is automatically satisfied, even though the Markov chain might diverge. A simple example is given by $A=2$, $B=1$, $T=0$, $U=0$, $R=1$, for any $\gamma\in ]0,1]$. The system is unstable since $K=0$ but $V_\star^\gamma (\vect{s})=0$, such that $\lim_{k\to\infty} V_\star^\gamma (\vect{s})=0$ holds. In this case we have $\tilde L^\gamma(\vect{s},\vect{a}) = L(\vect{s},\vect{a})$. 

\paragraph*{A Simple Example}
\label{sec:analytic_lqr}

Consider a linear system of the form~\eqref{eq:linear_mdp} with $A=2$, $B=1$, $T=1$, $U=0$, $R=1$, and set $W=0$ in order to have a deterministic system. One can verify that this system is stabilized only for feedback matrices $K\in \mathcal{K} := \ ]1,3[$, since this implies $A-BK\in \ ]-1,1[$.
The discounted LQR solution is 
\begin{subequations}
	\begin{align}
		P &= \frac{5\gamma-1 + \sqrt{ (1-5\gamma)^2 +4\gamma }}{2\gamma}, \\
		K &= \frac{4\gamma}{1-3\gamma + \sqrt{ (1-5\gamma)^2 +4\gamma }}, \label{eq:example_K}
	\end{align}
\end{subequations}
such that $\gamma\in \ ]1/3,1] \implies K\in\mathcal{K}$.

\begin{figure}
	\includegraphics[width=\linewidth]{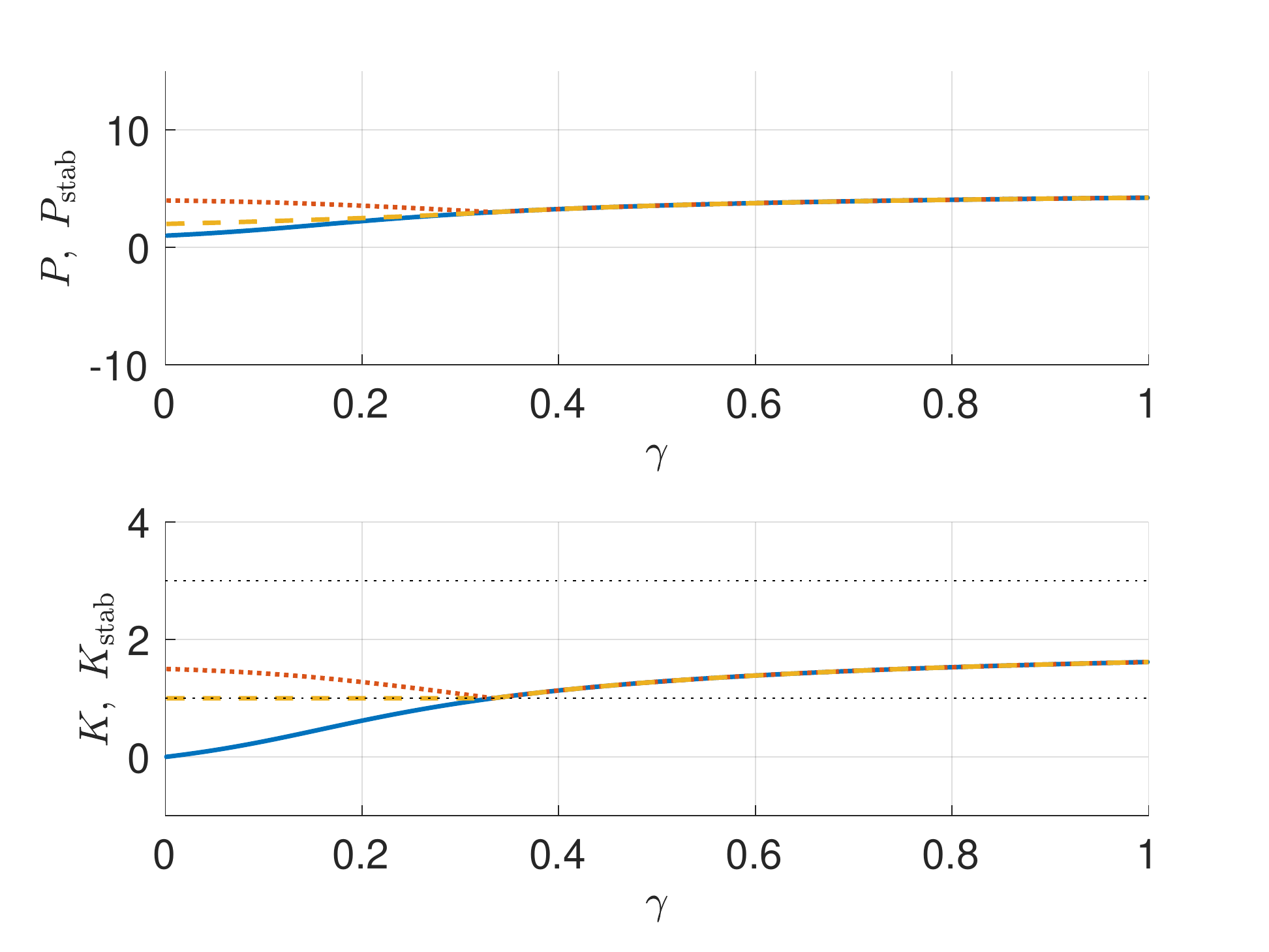}
	\caption{Analytic solution of the LQR problem: unconstrained solution (blue continuous line), stability constrained solution (yellow dashed line), solution with $\tilde L$ and no terminal cost (dotted red line), and stability bounds (black dotted lines).}
	\label{fig:analytic_lqr}
\end{figure}

We can write the stability-constrained MDP problem as 
\begin{equation}
	\label{eq:example_K_stab}
	K_\mathrm{stab}:=\arg\min_K \ P_{K}(K) \qquad \mathrm{s.t.} \ K\in\mathcal{K},
\end{equation}
where $V_{-Ks}^\gamma (\vect{s})=P_{K}(K)\vect{s}^2$ and 
\begin{equation*}
	P_{K}(K) := \frac{K^2 + 1}{1-\gamma(2-K)^2}.
\end{equation*}
We ought to stress that~\eqref{eq:example_K_stab} is not well-posed, since its constraint set is open and a solution might not exist. In practice, one usually defines a closed subset of the open set $\mathcal{K}$ in order to (a) make the problem well-posed; and (b) avoid being too close to the stability margin and have some robustness to numerical errors. 
Since the closure $\mathrm{cl}\left (\mathcal{K}\right ) := [1,3]$ of $\mathcal{K}$ guarantees marginal stability of the closed-loop system, we discuss the solution of
\begin{equation*}
	K_\mathrm{mstab}:=\arg\min_K \ P_{K}(K) \qquad \mathrm{s.t.} \ K\in\mathrm{cl}\left (\mathcal{K}\right ),
\end{equation*}
One can verify that, using $K$ given by~\eqref{eq:example_K},
\begin{equation*}
	K_\mathrm{mstab} = \left \{
	\begin{array}{ll}
		1 &\gamma\leq 1/3 \\
		K & \gamma> 1/3
	\end{array}
	\right ..
\end{equation*}
The solution is shown in Figure~\ref{fig:analytic_lqr}.
The undiscounted equivalent problem is obtained by computing stage cost $\tilde L^\gamma$ as per~\eqref{eq:tildeL:lqr}. Note that, even though $\tilde L^\gamma$ is defined for all $\gamma$ provided that $V_\star^\gamma$ is bounded,  Theorem~\ref{thm:undiscounted_inf_hor} applies only if Assumption~\ref{ass:Vstability_inf} holds, which is not the case for $\gamma<1/3$. Indeed, if $\gamma<1/3$ the closed-loop system becomes unstable and $\lim_{k\to\infty} V_\star^\gamma (\vect{s}_k) = \infty$. In turn, this entails that in such cases the stability-constrained discounted MDP with stage cost $L$ does not have the same solution as the stability-constrained undiscounted MDP with stage cost $\tilde L^\gamma$. This fact can be observed in Figure~\ref{fig:analytic_lqr}. We stress that if we formulate the undiscounted MDP with stage cost $\tilde L^\gamma$ and terminal cost $V_\star^\gamma (\vect{s})$, then Theorem~\ref{thm:undiscounted} applies and the equivalence between the two MDPs holds for all $\gamma$. This situation is captured by the discrete algebraic Riccati equation, which has two solutions in this case, corresponding to the two MDP solutions: with and without the terminal cost $V_\star^\gamma $.

Finally, we ought to stress that, as well-known, if we consider the same system with nonzero covariance $W\neq0$,
 the optimal feedback coincides with the one of its deterministic counterpart, i.e., when $W=0$. 


\section{Stability Constraints based on MPC}
\label{sec:constraints}

In the previous section, we proved that a discounted MDP can be reformulated as an undiscounted MDP and we mentioned that this fact can be useful to introduce stability constraints in the MDP formulation. However, we did not discuss how this can be done in the general case. 
In this section, we exploit the theoretical results of this paper in order to propose a solution method for stability-constrained MDPs which can be implemented in practice. 

In order to solve an MDP, one must compute either the value function $V_\star^\gamma$, the policy $\vect{\pi}^\gamma_\star$, the action-value function $Q_\star^\gamma$ or a combination of these. Since their functional form is not known a priori and can be rather complicated, solving MDPs exactly is a notoriously difficult task. 
Therefore, practical approaches typically rely on some parametric function approximation $V_\vect{\theta}$, $\vect{\pi}_\vect{\theta}$, and $Q_\vect{\theta}$, where $\vect{\theta}$ denotes a set of parameter adjusting the function approximations, in order to make the problem tractable~\cite{Bertsekas2007,Bertsekas1996a}. The stability-constrained discounted MDP~\eqref{eq:stab_constr_disc_mdp} can then be formulated using function approximation as
	\begin{align}
		\label{eq:stab_constr_disc_mdp_fa}
		\min_{\vect{\theta}\in\Theta_\mathrm{s}} \ \ \E{{\vect{\pi}_\vect{\theta}}}{\sum_{k=0}^{\infty} \gamma^k L(\vect{s}_k,\vect{\pi}_\vect{\theta}(\vect{s}_k)) },
	\end{align}
	where $\Theta_\mathrm{s}:=\{ \ \vect{\theta} \ | \ \vect{\pi}_\vect{\theta}\in\Pi_\mathrm{s} \ \}$ is the set of parameters which yield a stabilizing policy.
We propose to introduce the stability constraint $\vect{\theta}\in\Theta_\mathrm{s}$ by selecting an ad-hoc parametrization of these functions defining the MDP solution, which yields stabilizing policies by construction. These parametrizations will be based on MPC and will rely on tools from control theory.

\subsection{MPC: a Structured Function Approximator}

In order to enforce closed-loop stability, we propose to rely on Model Predictive Control (MPC) to support the necessary function approximations. This approach has been proposed in~\cite{Gros2020} in the context of Reinforcement Learning. We recall next how MPC provides a very convenient way to support a parametric approximation of $V_\vect{\theta}$, $\vect{\pi}_\vect{\theta}$, and $Q_\vect{\theta}$, as it solves the optimal control problem
\begin{subequations}%
	\label{eq:param_nmpc}%
	\begin{align}%
	\hspace{-0.8em}Q_\vect{\theta}(\vect{s},\vect{a}) =  \min_{\vect{z}}\ & \lambda_\vect{\theta}(\vect{s})+  V^\mathrm{f}_\vect{\theta}(\vect{x}_N) + \hspace{-1pt}\sum_{k=0}^{N-1} \ell_\vect{\theta}(\vect{x}_k,\vect{u}_k) \label{eq:param_nmpc:cost}\\
	\mathrm{s.t.} \ & \vect{x}_0 = \vect{s}, \quad \vect{u}_0=\vect{a}, \label{eq:param_nmpc:initial}\\
	&\vect{x}_{k+1} = \vect{f}_\vect{\theta}\left(\vect{x}_k,\vect{u}_k\right), \label{eq:param_nmpc:dynamics}\\
	& \vect{g}\left(\vect{u}_k\right) \leq 0, \label{eq:param_nmpc:input_const} \\
	& \vect{h}_{\vect{\theta}}\left(\vect{x}_k,\vect{u}_k\right) \leq 0,\quad \vect{h}^\mathrm{f}_\vect{\theta}(\vect{x}_N) \leq 0, \label{eq:Const:Relaxation} 
	\end{align}
\end{subequations}
where $\vect{z}=(\vect{x}_0,\vect{u}_0,\ldots,\vect{x}_N)$. The stage and terminal cost $\ell_\vect{\theta}, V^\mathrm{f}_\vect{\theta}$, the system dynamics and constraints $\vect{f}_\vect{\theta},\vect{h}_\vect{\theta},\vect{h}^\mathrm{f}_\vect{\theta}$ and the initial cost $\lambda_\vect{\theta}$ are all parametric functions of $\vect{\theta}$, while actuator limitations $\vect{g}$ are  known. Note that in MPC the initial constraint~\eqref{eq:param_nmpc:initial} typically only involves the state, i.e., $\vect{u}_0=\vect{a}$ is not present, since the goal is to compute an optimal policy. The policy $\vect{\pi}_\vect{\theta}(\vect{s})$ and value function $V_\vect{\theta}(\vect{s})$ are obtained by solving Problem~\eqref{eq:param_nmpc} with constraint $\vect{u}_0=\vect{a}$ removed. This is fully equivalent to
\begin{equation}
\label{eq:param_nmpc_s}
\hspace{-0.5em} \vect{\pi}_\vect{\theta}(\vect{s}) = \mathrm{arg}\min_\vect{a}\, Q_\vect{\theta}(\vect{s},\vect{a}), \quad V_\vect{\theta}(\vect{s}) = \min_\vect{a}\, Q_\vect{\theta}(\vect{s},\vect{a}).
\end{equation}

We briefly comment on this particular MPC formulation. 
Function $\lambda_\vect{\theta}$ has been introduced in~\cite{Gros2020} to make it possible to use a positive-definite stage cost $\ell_\vect{\theta}$ even when the true stage cost $l$ is not. This choice is related to the stability theory of economic MPC, where $\lambda$ is called a \emph{storage function}. In~\cite{Gros2020} it is discussed in detail how the use of a parametrized stage cost $\ell_\vect{\theta}$ makes it possible to recover the optimal policy and value functions using (36)-(37) even if the MPC model \eqref{eq:param_nmpc:dynamics} does not accurately capture the system dynamics \eqref{eq:TrueDynamics}.


We ought to stress that, though we formulated~\eqref{eq:param_nmpc} using a notation which is easily interpreted as a deterministic formulation, any MPC formulation can be used, including stochastic and robust formulations. Note also that in some cases, the scheme is reformulated using time-varying constraints, e.g., in case of tube-based robust MPC (used in the safety-constrained MDP context in~\cite{Zanon2021}), where the constraints are tightened in order to guarantee that the original constraints are satisfied for all possible perturbations acting on a nominal model of the system. We stress that this is purely a matter of notation and implementation, and~\eqref{eq:param_nmpc} covers this case, provided that $\vect{f}_\vect{\theta}$ yields a set-valued state propagation.


After having recalled that MPC can be used as a function approximator when solving MDPs, we now discuss how stability can be enforced directly in MPC, such that the obtained MDP solution must also be stabilizing by construction. There exists a plethora of MPC formulations relying on different assumptions and providing different stability and safety guarantees and we refer to, e.g.,~\cite{Rawlings2017,Grune2011,Mayne2014,Mayne2018} and references therein for an overview. 
In this paper, for brevity we focus on two special cases: nominal and tube-based robust MPC. 


\paragraph*{Nominal MPC.}
Nominal asymptotic stability builds on the assumption that the deterministic model used in MPC is exact. This is clearly a simplified setting and we will consider a more general approach with robust MPC.

\begin{Assumption}
	\label{ass:mpc1}
	The system dynamics and stage cost are continuous at $(\vect{s}_\mathrm{s},\vect{a}_\mathrm{s})$ and satisfy $\vect{f}_\vect{\theta}(\vect{s}_\mathrm{s},\vect{a}_\mathrm{s})=\vect{s}_\mathrm{s}$, $\ell_\vect{\theta}(\vect{s}_\mathrm{s},\vect{a}_\mathrm{s})=0$. Moreover,
	$\ell_\vect{\theta}(\vect{s},\vect{a})\geq \alpha_1(\|\vect{s}-\vect{s}_\mathrm{s}\|)$, for some $\mathcal{K}_\infty$ function $\alpha_1$.
\end{Assumption}
\begin{Assumption}
	\label{ass:mpc2}
	There exists a terminal control law $\vect{\kappa}_\vect{\theta}(\vect{s})$ such that $\forall \ \vect{s} \ \mathrm{s.t.} \ \vect{h}^\mathrm{f}_\vect{\theta}(\vect{s}) \leq 0$, it holds that 
	\begin{align*}
		\vect{h}^\mathrm{f}_\vect{\theta}(\vect{f}_\vect{\theta}(\vect{s},\vect{\kappa}(\vect{s}))) &\leq 0, \\
		\vect{g}\left(\vect{\kappa}_\vect{\theta}(\vect{s})\right) &\leq 0, &&  \vect{h}_{\vect{\theta}}\left(\vect{s},\vect{\kappa}_\vect{\theta}(\vect{s})\right) \leq 0,
	\end{align*} 
	and the terminal cost satisfies
	\begin{equation*}
		V_\vect{\theta}^\mathrm{f}(\vect{s}) \geq V_\vect{\theta}^\mathrm{f}(\vect{f}_\vect{\theta}(\vect{s},\vect{\kappa}(\vect{s}))) + \ell_\vect{\theta}(\vect{s},\vect{\kappa}_\vect{\theta}(\vect{s}))).
	\end{equation*}
	Moreover, $\vect{h}^\mathrm{f}_\vect{\theta}(\vect{s}_\mathrm{s}) < 0$, and $V_\vect{\theta}^\mathrm{f}(\vect{s}) \leq \alpha_2(\|\vect{s}-\vect{s}_\mathrm{s}\|)$, for some $\mathcal{K}_\infty$ function $\alpha_2$.
\end{Assumption}

\begin{Proposition}
	\label{prop:mpc_stability}
	Suppose that the system~\eqref{eq:TrueDynamics} matches the MPC model $\vect{f}_\vect{\theta}$, and Assumptions~\ref{ass:mpc1} and~\ref{ass:mpc2} hold. Then, if MPC is feasible at the initial time instant, it will remain feasible at all future times and the closed-loop system is asymptotically stable.
\end{Proposition}
\begin{pf}
	This standard result is given in, e.g.,~\cite{Rawlings2017,Grune2011}.
\end{pf}


The introduction of stability constraints can then be done based on the result given in Proposition~\ref{prop:mpc_stability}: in order for Assumptions~\ref{ass:mpc1}-\ref{ass:mpc2} to hold, we select a parametrization which is smooth enough, define functions $\alpha_1$, $\alpha_2$ and constant $\epsilon$ to define the set of feasible parameters as
\begin{align}
	\Theta := \{ \ \vect{\theta} \ | \ & \ell_\vect{\theta}(\vect{s},\vect{a})\geq \alpha_1(\|\vect{s}-\vect{s}_\mathrm{s}\|), \nonumber \\
	&V_\vect{\theta}^\mathrm{f}(\vect{s}) \leq \alpha_2(\|\vect{s}-\vect{s}_\mathrm{s}\|), \nonumber \\
	&\vect{h}^\mathrm{f}_\vect{\theta}(\vect{s}_\mathrm{s}) \leq \epsilon, \ \vect{s}_\mathrm{s}=\vect{f}_\vect{\theta}(\vect{s}_\mathrm{s},\vect{a}_\mathrm{s}), \nonumber \\
	&\vect{h}^\mathrm{f}_\vect{\theta}(\vect{f}_\vect{\theta}(\vect{s},\vect{\kappa}_\vect{\theta}(\vect{s}))) \leq 0, \nonumber \\
	& \vect{g}\left(\vect{\kappa}_\vect{\theta}(\vect{s})\right) \leq 0,\  \vect{h}_{\vect{\theta}}\left(\vect{s},\vect{\kappa}_\vect{\theta}(\vect{s})\right) \leq 0 \nonumber \\
	& V_\vect{\theta}^\mathrm{f}(x) \geq V_\vect{\theta}^\mathrm{f}(\vect{f}_\vect{\theta}(\vect{s},\vect{\kappa}_\vect{\theta}(\vect{s}))) + \ell_\vect{\theta}(\vect{s},\vect{\kappa}_\vect{\theta}(\vect{s}))), \nonumber \\
	& \forall \ \vect{s} \ \mathrm{s.t.} \ \vect{h}^\mathrm{f}_\vect{\theta}(\vect{s}) \leq 0 \ \}.
	\label{eq:nominal_stab_set}
\end{align}
While this formulation might seem rather abstract and difficult to implement, in practice it is common to select $\ell_\vect{\theta}$ and $V_\vect{\theta}^\mathrm{f}$ as convex functions, for which it is simpler to impose upper and lower bounds. The steady-state constraint $\vect{s}_\mathrm{s}=\vect{f}_\vect{\theta}(\vect{s}_\mathrm{s},\vect{a}_\mathrm{s})$ might be necessary if either the model or the steady state depend on $\vect{\theta}$, unless one relies on the results of~\cite{Rawlings2008a} for infeasible setpoint tracking. Constraint $\vect{h}^\mathrm{f}_\vect{\theta}(\vect{s}_\mathrm{s}) \leq \epsilon$ might not be necessary, provided that some weak controllability assumption holds~\cite{Rawlings2017}. Finally, MPC formulations with a terminal point constraint only require the lower bound on the stage cost, provided that a weak controllability assumption holds.


Since this formulation only provides formal stability guarantees under the strong assumption that the deterministic MPC model is exact, we discuss next the case of robust MPC, which formally guarantees that the real system will be asymptotically stabilized to a set.


\paragraph*{Robust MPC.}
In tube-based robust MPC, model mismatch is handled as additive bounded process noise. 
Since the nonlinear case can be computationally demanding, we focus on the linear case.
For a detailed discussion on the use of Linear tube-based Robust MPC (LRMPC) in the context of safety-constrained MDPs see~\cite{Zanon2021}. In the following, we briefly recall the problem formulation and the related stability and safety constraints. 
LRMPC is based on repeatedly solving
\begin{subequations} %
	\label{eq:robust_mpc}%
	\begin{align}%
	Q_{\vect{\theta}}(\vect{s},\vect{a}):= \hspace{-1em}& \nonumber\\
	\min_{\vect{z}} \ \ & \sum_{k=0}^{N-1} \norm{\vect{x}_k - \vect{x}_\mathrm{r} \\ \vect{u}_k  - \vect{u}_\mathrm{r}  }^2_H 
	+ \norm{\vect{x}_N - \vect{x}_\mathrm{r}}^2_P 
	\nonumber \\
	&\hspace{7em}
	+ \norm{\vect{x}_0}^2_\Lambda + \vect{\lambda}^\top \vect{x}_0  + l
	\label{eq:robust_mpc_cost} \hspace{-10em}&\hspace{0em}\\ 
	\mathrm{s.t.} \ \ & \vect{x}_0 = \vect{s}, \qquad \vect{u}_0 = \vect{a}, \label{eq:robust_mpc_ic}\\
	& \vect{x}_{k+1} = A\vect{x}_k + B \vect{u}_k + \vect{b}, & \hspace{-3em} k\in\mathbb{I}_0^{N-1}, \label{eq:robust_mpc_dyn}\\
	& C\vect{x}_k + D \vect{u}_k + \vect{c}_k \leq \vect{0}, & \hspace{-3em} k\in\mathbb{I}_0^{N-1}, \label{eq:robust_mpc_pc}\\
	& T\vect{x}_N  + \vect{t} \leq \vect{0},\label{eq:robust_mpc_tc}
	\end{align}%
\end{subequations}%
where one must enforce that the system dynamics~\eqref{eq:robust_mpc_dyn} and a parametrized compact uncertainty set $\mathbb{W}_\vect{\omega}$ satisfy
\begin{equation*}
	\vect{s}_+ - (A\vect{s}+B\vect{a} + \vect{b}) \in \mathbb{W}_\vect{\omega}.
\end{equation*}
The set is typically parametrized as the polyhedron $\mathbb{W}_\vect{\omega}:=\{ \ \vect{w} \ | \ M\vect{w} \leq \vect{m} \ \}$ and the following set membership constraint is imposed on $M,\vect{m}$ for all past samples $\vect{s}_{i+1}, \vect{s}_i, \vect{a}_i$, $i\in\mathcal{I}$:
\begin{equation*}
	M(\vect{s}_{i+1} - (A\vect{s}_i+B\vect{a}_i + \vect{b})) \leq \vect{m}, \qquad \forall \ i\in\mathcal{I}.
\end{equation*}
Then, $\vect{c}_k$ is given by tightening the original constraints
\begin{equation*}
	C\vect{s} + D\vect{a} + \vect{\hat c} \leq 0,
\end{equation*}
so as to guarantee that, for any process noise $\vect{w}\in \mathbb{W}_\vect{\omega}$, the original constraints are satisfied. 
Parameters $\vect{x}_\mathrm{r},\vect{u}_\mathrm{r}$ must be a steady-state for the system dynamics~\eqref{eq:robust_mpc_dyn}:
\begin{equation*}
	(A-I) \vect{x}_\mathrm{r} + B \vect{u}_\mathrm{r} = \vect{0}.
\end{equation*}
Finally, $T$ and $\vect{t}$ must be selected such that they define a robust positively invariant terminal set for the feedback law $\vect{u}=-K(\vect{x}-\vect{x}_\mathrm{r}) + \vect{u}_\mathrm{r}$, with $K$ the solution to the LQR formulated with $A,B,H,P$. 
The vector of MPC parameters is then defined as
\begin{equation}
	\vect{\theta} = \{ \Lambda, \lambda, l, H, \vect{x}_\mathrm{r}, \vect{u}_\mathrm{r}, M \},
\end{equation}
and we consider $K$, $P$, $\vect{c}_k$, $T$, $\vect{t}$ as functions of these parameters. Vector $\vect{m}$ can also be included in $\vect{\theta}$, but, as discussed in~\cite{Zanon2021} this is not necessary. Matrices $C$, $D$ and vector $\vect{\bar{c}}$ are assumed to be known. Finally, while $A$, $B$, $\vect{b}$ can in principle also be included in the parameter vector $\vect{\theta}$. However, as discussed in~\cite{Zanon2021} modifying $A$ and $B$ makes the MDP much harder to formulate and solve.

\begin{Proposition}
	Assume that the terminal cost is selected as the solution of the Riccati equation and the corresponding feedback law is used both to predict the uncertainty evolution when performing constraint tightening and to define the terminal positive invariant set. 
	Assume moreover that, if the nominal prediction satisfies the tightened path and terminal constraints, then the real system satisfies the original path constraints and remains in the positive invariant terminal set.  Then, the true system is asymptotically stabilized to the minimum robust positive invariant set associated with the Riccati feedback.
\end{Proposition}
\begin{pf}
	This result can be found in, e.g.,~\cite{Brunner2018,Zanon2021a}.
\end{pf}

For some small $\epsilon>0$, the set of parameters guaranteeing safety and stability then becomes
\begin{align*}
\Theta := \{ \ \vect{\theta} \ | \ & H\succeq \epsilon I, \\
& M(\vect{s}_{i+1} - (A\vect{s}_i+B\vect{a}_i + \vect{b})) \leq \vect{m}, \ \forall \ i \in \mathcal{I}, \\
& (A-I) \vect{x}_\mathrm{r} + B \vect{u}_\mathrm{r} = \vect{0}, \\
&\exists \ \vect{x} \ \mathrm{s.t.} \ T\vect{x} \leq \vect{t}
 \ \},
\end{align*}
i.e., the noise set must include all observed noise samples, the reference must be a steady-state of the system and the terminal set must be nonempty. This last condition also entails that the MPC domain is nonempty.

\subsection{MPC Approximator and MDP Optimality}

We will prove next that undiscounted MPC can yield a function approximator delivering the exact solution of the discounted MDP, even if the MPC model is not exact, e.g., if MPC relies on a deterministic model while the true system~\eqref{eq:TrueDynamics} is stochastic. To that end, we will first recall a result from~\cite{Gros2020} which establishes the equivalence in case MPC has the same discount factor as the MDP. We will exploit the equivalence with $\gamma=1$ to prove that MPC can support the exact solution of the undiscounted MDP. The desired result is then obtained by exploiting the equivalence between the undiscounted and the discounted MDP, stated in Theorem~\ref{thm:undiscounted_inf_hor}.

We recall next the main result of~\cite{Gros2020}, adapted to the context of interest in this paper. 
Consider an undiscounted MDP with a given stage cost\footnote{This stage cost need not be related to $\tilde L^\gamma$ for the results of~\cite{Gros2020}.} $\tilde L$, zero optimal gain, optimal action-value and value function $\tilde Q_\star$, $\tilde V_\star$ and bias-optimal policy $\vect{\tilde \pi}_\star$. Define the stage cost
\begin{align*}
	\hat L(\vect{s},\vect{a}) &:= \left \{  
	\begin{array}{ll}
		\tilde Q_\star(\vect{s},\vect{a}) - \mathcal{\tilde V}^+(\vect{s},\vect{a}) & \text{if} \ \left | \mathcal{\tilde V}^+(\vect{s},\vect{a})  \right | < \infty \\
		\infty & \text{otherwise}
	\end{array}
	\right . , \\
	\mathcal{\tilde V}^+(\vect{s},\vect{a}) &:= \E{}{\tilde V_\star\left (\vect{f}_\vect{\theta}(\vect{s},\vect{a})\right )}.
\end{align*}
Note that $\hat L=\tilde L$ only in case $\vect{f}_\vect{\theta}(\vect{s},\vect{a})$ yields the exact system dynamics~\eqref{eq:TrueDynamics}. Further denote as $\hat V_\star$ the value function associated with solving
\begin{align*}
	\min_{\vect\pi}\,  \lim_{N\to\infty} \E{\vect{\pi}}{ \ \sum_{k=0}^{N-1} \hat L(\vect{s}_k,\vect{\pi}(\vect{s}_k)) \   }.
\end{align*}
\begin{theorem}[\cite{Gros2020}]
	\label{thm:wrong_model}
	Consider an undiscounted MDP with stage cost $\tilde L$, optimal value function $\tilde V_\star$ and bias-optimal policy $\vect{\tilde \pi}_\star$, such that under policy $\vect{\tilde \pi}_\star$ 
	\begin{align}
		\label{eq:stability_assumption_replicated}
		&\lim_{N\to\infty}\tilde V_\star(\vect{s}_N)=0, &&&
		& \left |\tilde V_\star(\vect{s}_k)\right |<\infty, \quad \forall \ k \geq 0.
	\end{align}
	Assume that the parametrization of the MPC scheme~\eqref{eq:param_nmpc} is descriptive enough, i.e., $\exists \, \vect{\theta}_\mathrm{o}\in\Theta$ such that
	\begin{subequations}
		\label{eq:exact_approx}
		\begin{align}
			&\ell_{\vect{\theta}_\mathrm{o}}(\vect{s},\vect{a}) = \hat L (\vect{s},\vect{a}), &&\hspace{-0.5em}\forall  \ \vect{s},\vect{a} \, \mathrm{s.t.} \, \vect{h}_{\vect{\theta}_\mathrm{o}}\left(\vect{s},\vect{a}\right)\leq0, \\
			&V_{\vect{\theta}_\mathrm{o}}^\mathrm{f}(\vect{s}) = \hat V_\star(\vect{s}), &&\hspace{-0.5em}\forall \, \vect{s} \, \mathrm{s.t.} \, \vect{h}_{\vect{\theta}_\mathrm{o}}^\mathrm{f}\left(\vect{s}\right)\leq0, \\
			&\left |\hat V_\star(\vect{s})\right | <\infty \Longleftrightarrow \left |V_\vect{\theta}(\vect{s})\right | < \infty. \hspace{-10em} 
		\end{align}
	\end{subequations}
	Then, using MPC~\eqref{eq:param_nmpc} as function approximator, the exact optimal value function and policy are recovered.
\end{theorem}
\begin{pf}
	The proof follows from~\cite[Theorem~1, Corollary~2]{Gros2020}. $\hfill\qed$
\end{pf}

We now apply this result to the undiscounted MDP built with stage cost $\tilde L^\gamma$ defined in~\eqref{eq:Ltilde:def}, in order to also obtain the equivalence for the original discounted MDP, thanks to Theorem~\ref{thm:undiscounted_inf_hor}.
\begin{theorem}
	\label{thm:exact_solution}
	Suppose that Assumption~\ref{ass:Vstability_inf} holds and the parametrization of MPC~\eqref{eq:param_nmpc} satisfies the approximation conditions~\eqref{eq:exact_approx} for $\hat L$ defined using $\tilde L^\gamma$ and the corresponding value functions.
	Then, the solution of the stability-constrained MDP~\eqref{eq:stab_constr_disc_mdp_fa} which supports the policy using MPC scheme~\eqref{eq:param_nmpc} delivers the exact solution to MDP~\eqref{eq:stab_constr_disc_mdp}.
\end{theorem}
\begin{pf}
%
	By Theorem~\ref{thm:wrong_model}, the exact solution of the undiscounted MDP with stage cost $\tilde L^\gamma$ can be recovered, provided that~\eqref{eq:stability_assumption_replicated} holds, which is a direct consequence of Assumption~\ref{ass:Vstability_inf} and Theorem~\ref{thm:undiscounted_inf_hor}. Moreover, by Theorem~\ref{thm:undiscounted_inf_hor} the solution of the undiscounted MDP with stage cost $\tilde L^\gamma$ matches the solution of the discounted MDP with stage cost $L$ and discount factor $\gamma$. $\hfill\qed$
\end{pf}

The result of Theorem~\ref{thm:exact_solution} guarantees that the optimal policy for~\eqref{eq:stab_constr_disc_mdp} is recovered. In turn, this entails that, if the original unconstrained MDP is already stabilizing, then the optimal policy is obtained. Otherwise, the stability constrained MDP will return, among all stabilizing policies, the one which minimizes the expected total discounted cost, since the unconstrained optimal one is not stabilizing.

Guaranteeing that the parametrization of MPC is rich enough to solve the MDP exactly is in general very difficult and the solution is approximate. This is due to the fact that $\hat L$, $\hat V_\star$, and the constraint functions can take rather convoluted forms. Additionally, in order to keep the MPC problem tractable, a simple parametrization is usually favored. Furthermore, the stability conditions for MPC are sufficient but typically not necessary, such that some stabilizing policies might be unnecessarily ruled out. 
A thorough analysis of the approximation properties of parametrizations which do not satisfy~\eqref{eq:exact_approx} is beyond the scope of this paper, but we observe that, by selecting a given parametrization, one selects the class of policies that can be supported and, consequently, the degree of suboptimality which can be achieved.

\section{Simulation Examples}
\label{sec:simulations}

In this section we provide three examples. The first two consider nominal asymptotic stability for: (a) a linear system with quadratic cost which diverges for small $\gamma$; and (b) for a nonlinear system for which steady-state operation is not optimal. Finally, we consider a stochastic system for which we provide safety and stability guarantees by robust MPC.

\subsection{Simple LQR}
\begin{figure}
	\includegraphics[width=\linewidth]{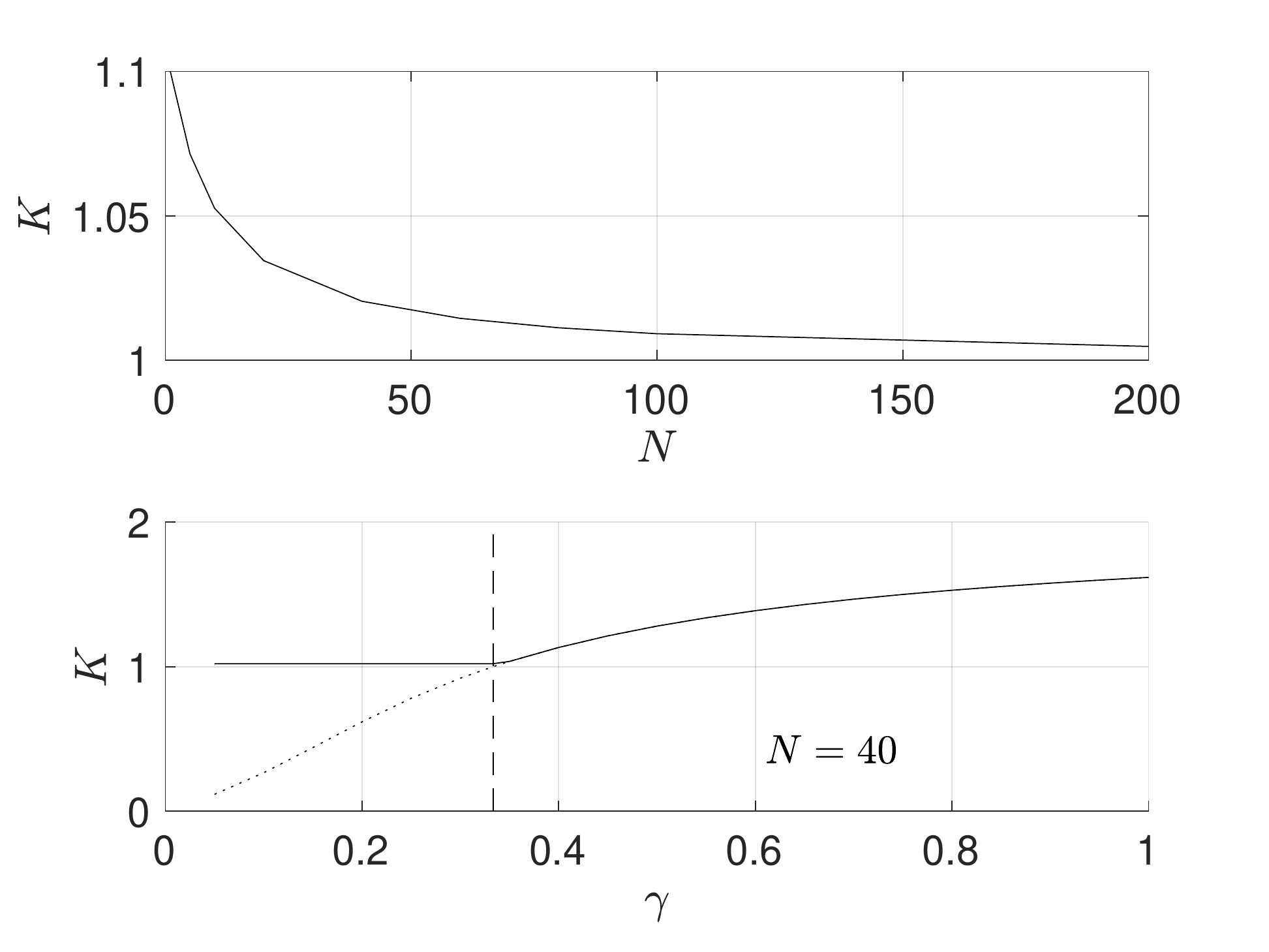}
	\caption{Batch Q-learning applied to the simple LQR example. Top plot: learned feedback for $\gamma=0.1$ and varying prediction horizon.
		Bottom plot: feedback learned with $N=40$ and safety constraints (continuous line) and unconstrained discounted LQR feedback (dotted line).
	}
	\label{fig:q_learning_lqr}
\end{figure}

We consider first the very simple example of Section~\ref{sec:analytic_lqr}. In order to simplify the interpretation of the results, we consider the case of no process noise.
We parametrize the problem by the initial, stage and terminal cost matrices $\Lambda$, $H$, $P$, where all costs are purely quadratic. We deploy Q-learning in $2000$ batches of $10$ samples; we use as terminal controller $\vect{a} = -K_\mathrm{f} \vect{s} = -1.2 \vect{s}$ and satisfy Assumption~\ref{ass:mpc1} by formulating~\eqref{eq:nominal_stab_set} with $\epsilon=10^{-8}$ as
\begin{equation*}
	H \succeq \epsilon I, \qquad 
	P \geq (A-BK_\mathrm{f})^2P + \tightMatr{c}{I\\ -K_\mathrm{f}}^\top H \tightMatr{c}{I\\ -K_\mathrm{f}}.
\end{equation*}
We initialize the parameters as $\Lambda=0$, $H=I$, $P=1$. We display the simulation results in Figure~\ref{fig:q_learning_lqr}. One can see that for $\gamma=0.1$ the RL feedback tends to $1$ as the prediction horizon increases. The fact that the feedback does not go to $1$ is due to the fact that in the terminal conditions we use $K_\mathrm{f}=1.2$. One possible remedy is to update the terminal feedback with the learned one. In that case, the feedback learned by RL is independent of the prediction horizon, but depends on the value of $\epsilon$ and on the numerical accuracy of the SDP solver.
By fixing $N=40$ and solving the problem for $\gamma\in]0,1]$, we observe that RL learns the correct feedback for all $\gamma$, i.e., it matches the discounted LQR feedback for all $\gamma>1/3$ and returns a feedback close to $1$ otherwise. This result coincides with the analytical derivations of Section~\ref{sec:analytic_lqr}.


\subsection{Nominal Nonlinear Economic MPC}

Consider the following Continuously Stirred Tank Reactor (CSTR) from~\cite{Diehl2011,Amrit2011a}, where a single irreversible chemical reaction $\mathrm{A}\to\mathrm{B}$ takes place with reaction rate $k_\mathrm{r}c_\mathrm{A}$, where $k_\mathrm{r}=0.4 \ \mathrm{l} / (\mathrm{mol} \, \mathrm{min})$ is the rate constant and $c_\mathrm{A}$, $c_\mathrm{B}$ are the concentrations of $\mathrm{A}$ and $\mathrm{B}$ respectively. The process dynamics are
\begin{equation*}
	\dot c_\mathrm{A} = \frac{q}{V_\mathrm{R}}(c_\mathrm{Af}-c_\mathrm{A})-k_\mathrm{r}c_\mathrm{A}, \qquad
	\dot c_\mathrm{B} = \frac{q}{V_\mathrm{R}}(c_\mathrm{Bf}-c_\mathrm{B})+k_\mathrm{r}c_\mathrm{A},
\end{equation*}
where $c_\mathrm{Af}=1 \ \mathrm{mol}/\mathrm{l}$,  $c_\mathrm{Bf}=0 \ \mathrm{mol}/\mathrm{l}$ are the feed concentrations of $\mathrm{A}$ and $\mathrm{B}$, $V_\mathrm{R}$ is the volume of the reactor. The flow $q$ through the reactor is the control variable, which is constrained in the interval $[0,20] \ \mathrm{l}/\mathrm{min}$. The system is discretized using a sampling time $t_\mathrm{s}=1 \ \mathrm{min}$.

The discount factor is $\gamma=0.9$ and the stage cost is
\begin{equation}
	\ell(\vect{s},\vect{a}) = 2qc_\mathrm{A} - 1.5q.
\end{equation}
As observed in~\cite{Amrit2011a}, even though $\vect{s}_\mathrm{s}=(0.5,0.5)$, $\vect{a}_\mathrm{s}=4$ is an economically optimal steady-state, this cost does not yield asymptotic stability to that steady state for $\gamma=1$, as periodic operation yields a lower cost. Also with $\gamma=0.9$ periodic operation does yield a lower cost than operating the system at the optimal steady-state. 
We observe that, to the best of our knowledge, a formal method to solve this problem with formal steady-state stability guarantees is currently not available. 

We formulate a nominal MPC scheme using the simple quadratic stage and initial cost
\begin{align*}
	\ell_{\vect{\theta}}(\vect{x},\vect{u}) &= \norm{\vect{x} - \vect{s}_\mathrm{s} \\ \vect{u}  - \vect{a}_\mathrm{s}  }^2_H, \\
	\lambda_\vect{\theta}(\vect{x})&=\norm{\vect{x} - \vect{s}_\mathrm{s}}^2_\Lambda + \vect{\lambda}^\top (\vect{x} - \vect{s}_\mathrm{s})  + l,
\end{align*}
with parameter vector $\vect{\theta} = (H,\Lambda,\vect{\lambda},l).$

\begin{figure}
	\includegraphics[width=\linewidth]{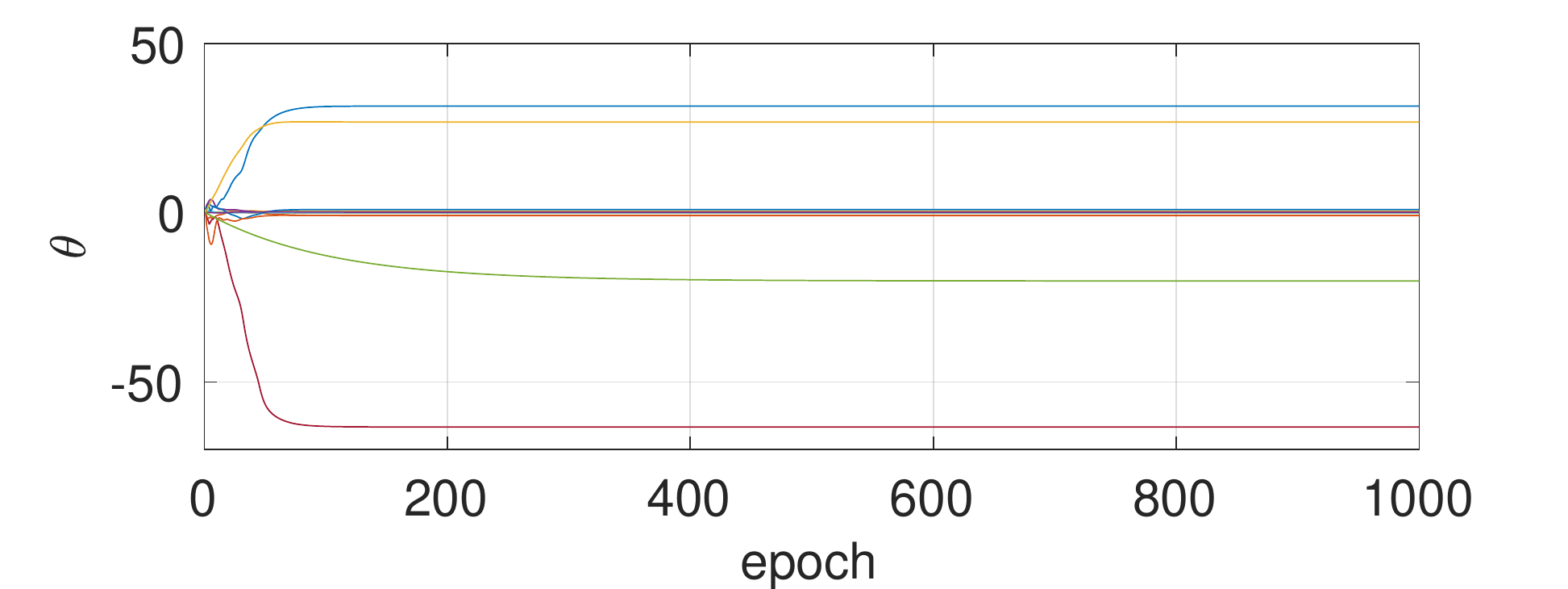}
%
	\includegraphics[width=\linewidth]{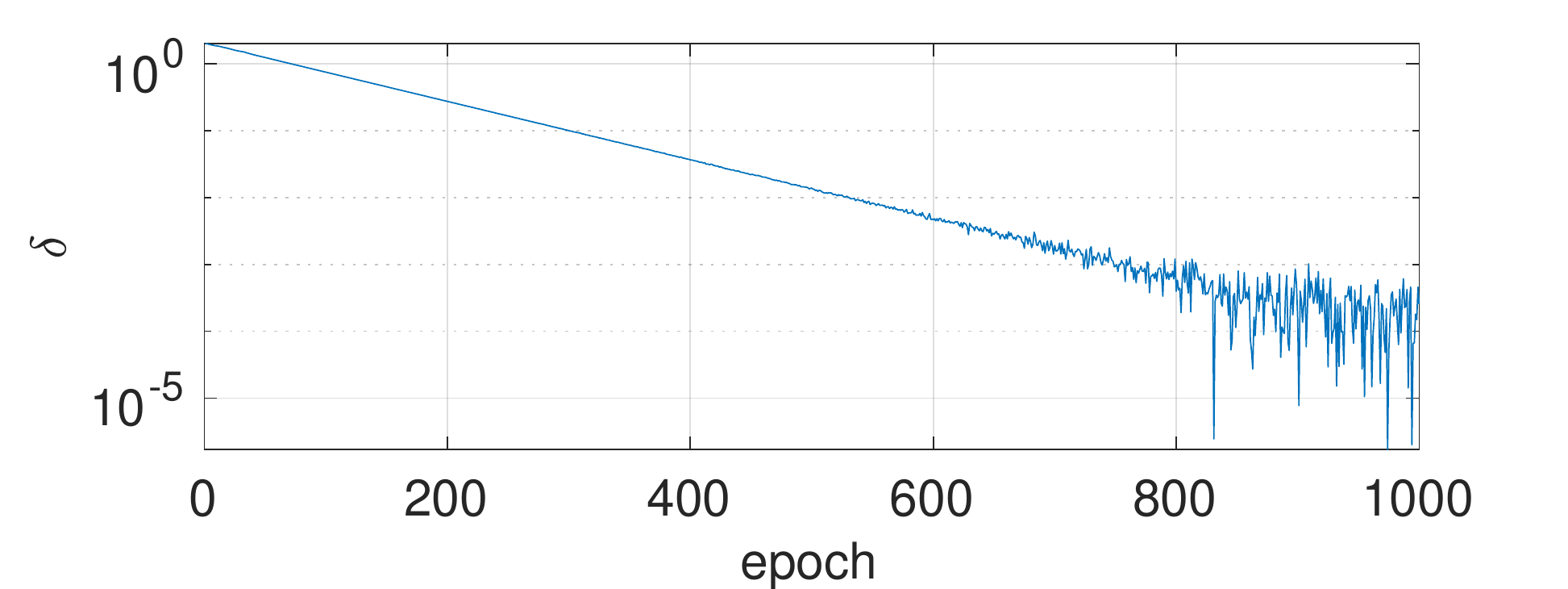}
	\caption{Evolution of the parameter $\vect{\theta}$ and the TD error $\delta$ over the RL epochs.}
	\label{fig:cstr_TDerror}
\end{figure}

In order to obtain simple conditions for asymptotic stability, we enforce a terminal point constraint $\vect{x}_N=\vect{s}_\mathrm{s}$, with a prediction horizon $N=100$. Since we enforce a terminal point constraint, a sufficient condition for nominal asymptotic stability is $H\succ \epsilon I$, with $\epsilon >0$. We select $\epsilon = 10^{-4}$ and run batch Q-learning with learning factor $\alpha = 0.1$ and $8$ batches of $40$ samples starting from initial conditions $(1,0)$, $(0,1)$, $(1,1)$, $(0,0)$, $(0.8,0.3)$, $(0.3,0.8)$, $(0.4,0.6)$, $(0.6,0.4)$. We run $1000$ epochs and obtain the parameter evolution and TD-error displayed in Figure~\ref{fig:cstr_TDerror} respectively. We observe that the obtained stage cost matrix has eigenvalues $31.5$, $10^{-4}$, $10^{-4}$, such that, as expected, the stability constraint is active.

\subsection{Robust MPC}

Consider the linear system with dynamics and stage cost
\begin{align*}
\vect{s}_+ &= \matr{cc}{1 & 0.1 \\ 0 & 1.1} \vect{s} + \matr{c}{0.05 \\ 0.1} \vect{a} + \vect{w}, \\
\ell(\vect{s},\vect{a}) &= \matr{c}{\vect{s}-\vect{s}^\mathrm{r} \\ \vect{a} -\vect{a}^\mathrm{r} }^\top \mathrm{diag}\left ( \matr{c}{1 \\[-0.5em] 0.01 \\[-0.5em] 0.01} \right ) \matr{c}{\vect{s}-\vect{s}^\mathrm{r} \\ \vect{a} -\vect{a}^\mathrm{r} },
\end{align*}
where $\vect{s}=(p,v)$ and $\vect{s}^\mathrm{r} =(-3,0)$, $\vect{a}^\mathrm{r}=0$, with discount factor  $\gamma = 0.5$. Note that the discounted unconstrained problem without stability constraints is not stabilizing.

We formulate tube based MPC as per~\eqref{eq:robust_mpc} with prediction horizon $N=50$ and introduce the state and control constraints $-\vect{1}\leq \vect{s}\leq \vect{1}$, $-10\leq \vect{a}\leq 10$.
The real noise set is selected as a regular octagon, and we parametrize $\mathbb{W}_{\vect{\omega}}$ as a polytope with 4 facets. 
We update $\vect{\theta}=\{\Lambda, \lambda, l, H, \vect{x}_\mathrm{r}, \vect{u}_\mathrm{r}, M\}$ using $Q$ learning with learning factor $\alpha=0.1$.

\begin{figure}
	\includegraphics[width=\linewidth]{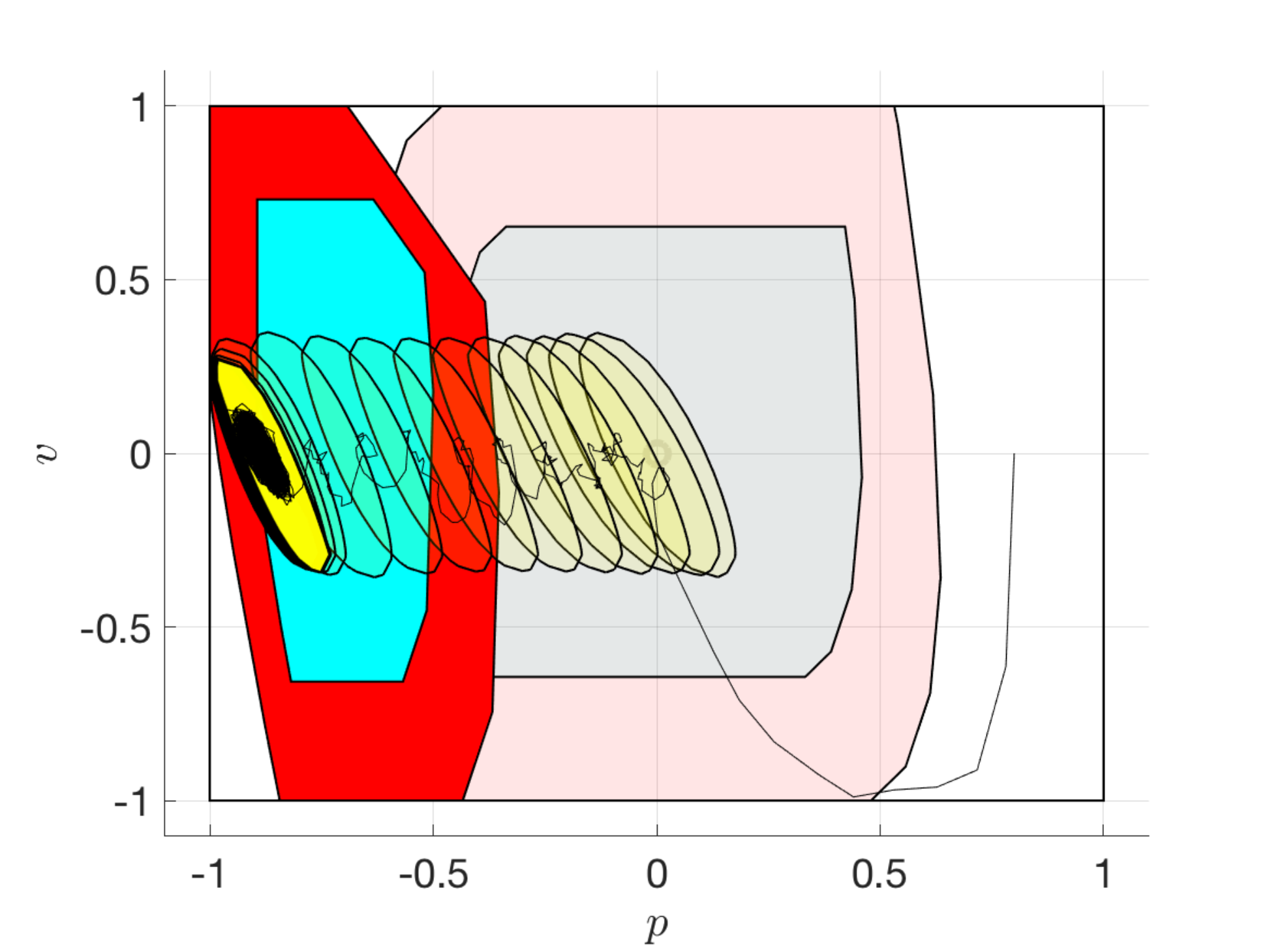}
	\caption{MRPI (red), terminal (cyan) sets and reference $\vect{x}^\mathrm{r}$ (black and grey circle) at the beginning and end of the learning process; state trajectory (black line) and mRPI sets (yellow) at each time instant.}
	\label{fig:tube_sets}
\end{figure}

\begin{figure}
	\includegraphics[width=\linewidth]{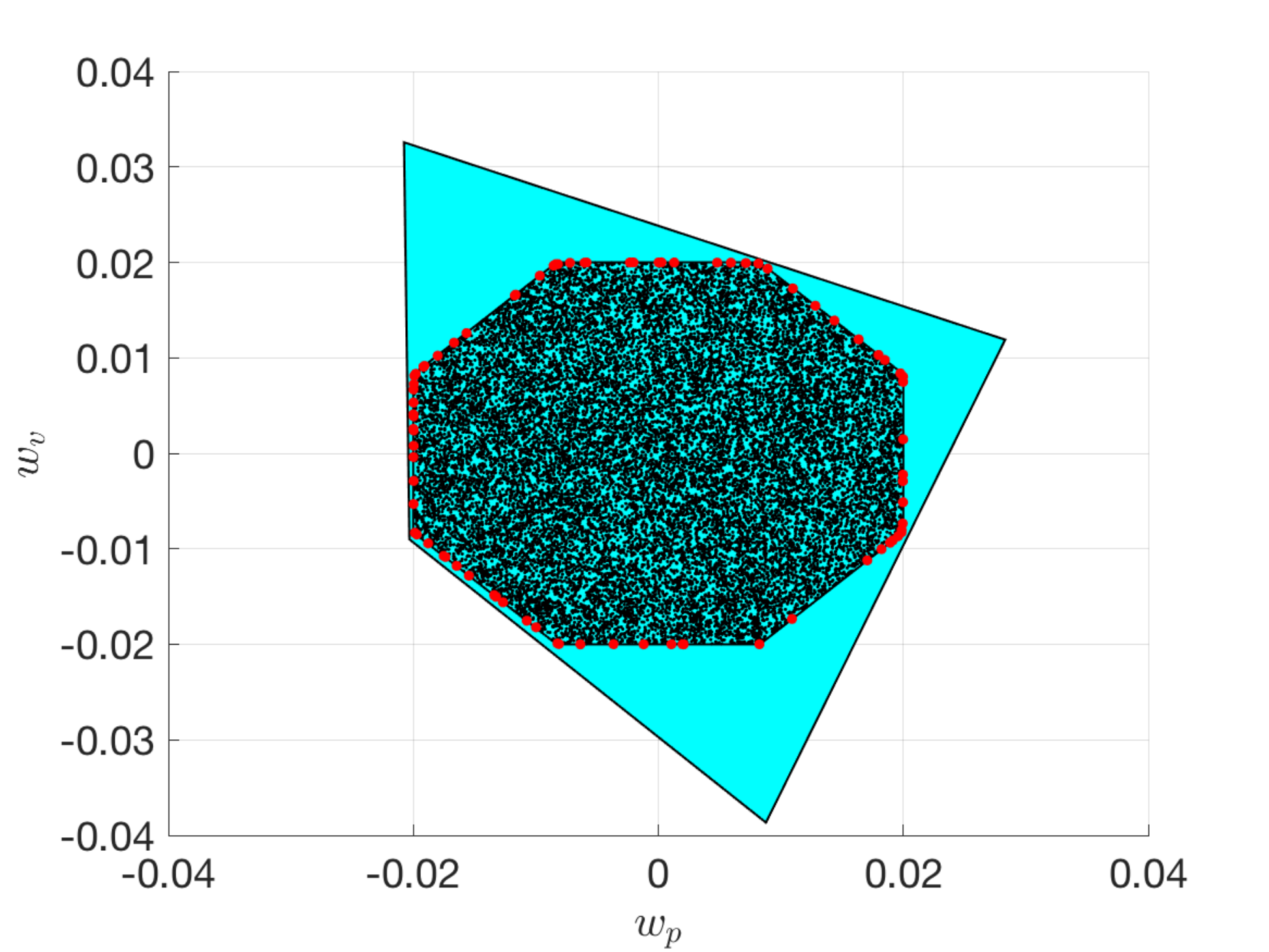}
	\caption{True process noise set (transparent octagon), noise samples (black dots), their convex hull (red dots) and noise set parametrized by matrix $M$ (cyan).}
	\label{fig:tube_noise}
\end{figure}

\begin{figure}
	\includegraphics[width=\linewidth]{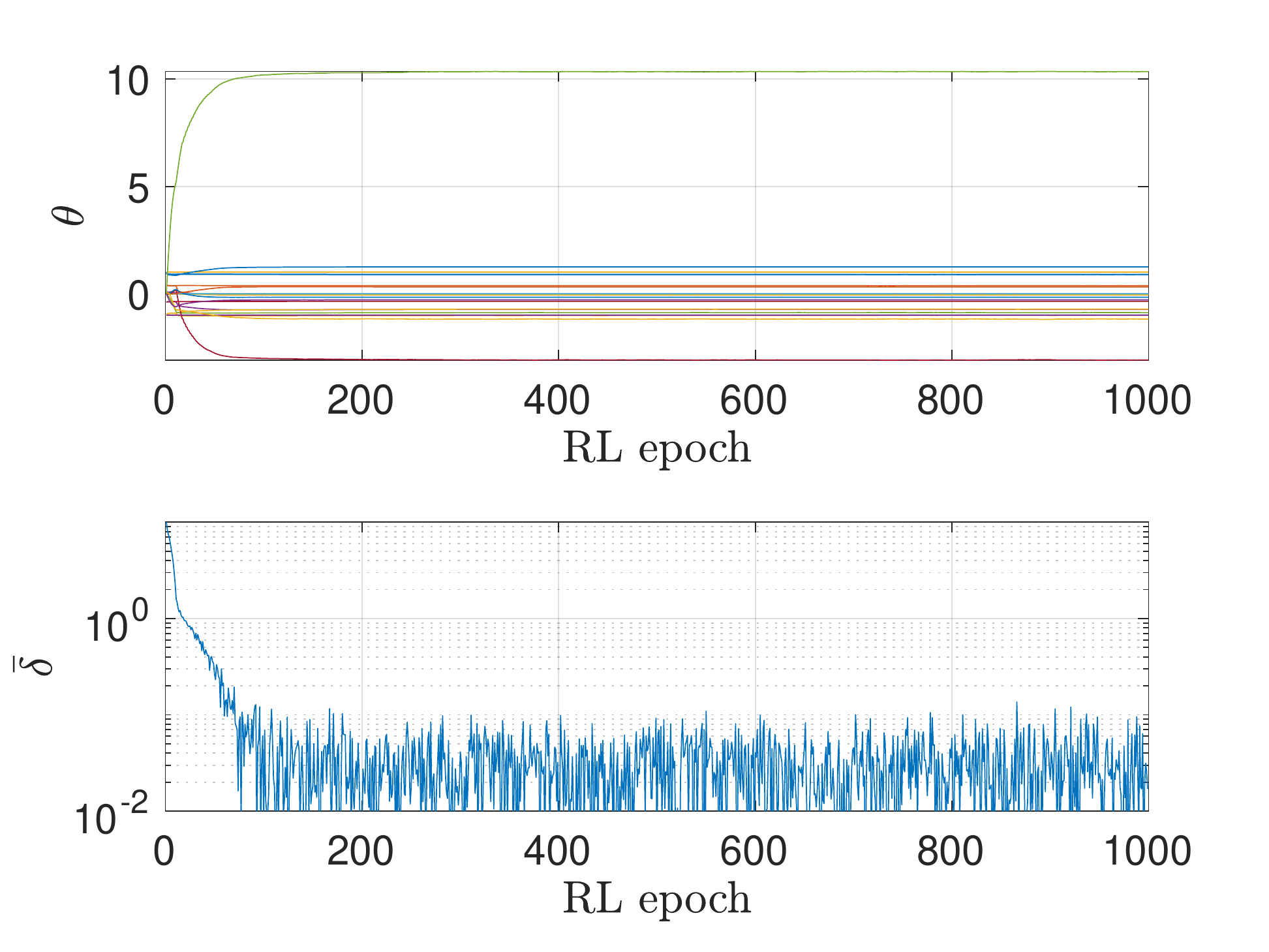}
	\caption{Top plot: parameter evolution through the epochs. Bottom plot: TD error through the epochs.}
	\label{fig:tube_params}
\end{figure}

The closed-loop trajectory starting from $\vect{s}_0=(0.8,0)$ is displayed in Figure~\ref{fig:tube_sets}, together with the reference, Maximum Robust Positive Invariant (MRPI) and terminal sets at the beginning and end of the simulation, as well as the minimum Robust Positive Invariant (mRPI) sets throughout the simulation. We display the noise set approximation at the end of the simulation in Figure~\ref{fig:tube_noise}, and the evolution throughout the RL epochs of the parameter $\vect{\theta}$ and the average TD error in each batch in Figure~\ref{fig:tube_params}.

\section{Conclusions}
\label{sec:conclusions}

In this paper, we have provided a way to enforce stability constraints in discounted MDPs. In order to achieve that, we have proven that, under a weak assumption, any discounted MDP can be reformulated as an undiscounted MDP, and we have proven that the obtained undiscounted MDP yields bias-optimal policies. In order to introduce the stability constraints, we exploited the results of~\cite{Gros2020} in order to deploy stabilizing MPC formulations to support the value functions and policy approximations required to solve the MDP. Future work will further investigate the stability properties of discounted MDPs and the possibility of using the equivalence in order to derive new algorithms for solving undiscounted MDPs.

\bibliographystyle{plain}
\bibliography{syscop}

\end{document}